\documentclass[11pt]{article}

\usepackage[preprint]{acl}

\usepackage{times}
\usepackage{latexsym}

\usepackage[T1]{fontenc}

\usepackage[utf8]{inputenc}

\usepackage{microtype}

\usepackage{inconsolata}

\usepackage{graphicx}

\usepackage{microtype}
\usepackage{hyperref}
\usepackage{url}
\usepackage{booktabs}
\usepackage{amsmath}

\usepackage{tikz}
\usepackage{booktabs}
\usepackage{multirow}

\usepackage{algorithm}
\usepackage{algcompatible}
\usepackage{algpseudocode}

\usepackage{lineno}

\definecolor{darkblue}{rgb}{0, 0, 0.5}
\hypersetup{colorlinks=true, citecolor=darkblue, linkcolor=darkblue, urlcolor=darkblue}

\usepackage{xcolor}
\usepackage[table]{xcolor}
\usepackage[most]{tcolorbox}
\usepackage[T1]{fontenc}

\usepackage{tcolorbox}
\tcbset{
  colback=gray!5,
  colframe=gray!70,
  boxrule=0.5pt,
  arc=3pt,
  left=6pt,
  right=6pt,
  top=6pt,
  bottom=6pt
}

\usepackage{tikz}

\newcommand{\circnum}[1]{%
\tikz[baseline=(char.base)]{
\node[
    shape=circle,
    draw,
    fill=blue!15,
    inner sep=2pt,
    font=\small
] (char) {#1};}}

\title{PARTAB: Partition-Aware Reasoning with Structured Evidence for Scalable Table Understanding}

\author{Md Mahadi Hasan Nahid \\
  University of Alberta\\
  \texttt{mnahid@ualberta.ca} \\\And
  Davood Rafiei \\
  University of Alberta \\
  \texttt{drafiei@ualberta.ca} \\}

\begin{document}
\maketitle




\begin{abstract}

Large Language Models (LLMs) have shown strong capabilities in table reasoning, but their effectiveness degrades as tables grow in size and complexity due to irrelevant context and difficulty localizing the evidence required for reasoning. Existing approaches typically reason over either the full table or a single reduced view, which can still obscure important row-column relationships. We introduce \textbf{PARTAB} (\textbf{P}artition-\textbf{A}ware \textbf{R}easoning over \textbf{Tab}les), a framework that constructs a structured evidence interface between the LLM and the table. PARTAB represents query-relevant evidence as semantically coherent, row-linked table regions and performs hierarchical selection over column groups and row-level partitions before composing the selected evidence for answer generation. We evaluate PARTAB on multiple table reasoning benchmarks, covering question answering, fact verification, and numerical reasoning. PARTAB consistently improves over full-table prompting and several recent table reasoning methods, achieving strong performance on WikiTableQuestions and TabFact while remaining competitive on numerical reasoning. Additional analyses show that semantic partitioning and targeted evidence selection improve evidence localization, substantially reduce the reasoning context, and provide larger benefits on complex tables. These results demonstrate the value of structured, partition-aware evidence construction for scalable table reasoning.

\end{abstract}

\section{Introduction}

Large Language Models (LLMs) have demonstrated strong performance on table reasoning tasks such as question answering and fact verification by operating directly over serialized tabular inputs~\citep{cheng2023binder,liu2022tapex}. However, their effectiveness degrades significantly as tables grow in size and complexity~\citep{wang2024chainoftable}. This raises a central research question: \emph{how can LLMs scale to large and noisy tables while accurately localizing and reasoning over the relevant evidence?}

A key challenge is that, in many table reasoning tasks, the required evidence is localized to a small set of regions, consisting of specific combinations of rows and columns, while existing approaches require reasoning over the entire table~\citep{nahid-rafiei-2024-tabsqlify}. As table size increases, irrelevant content introduces noise, leading to attention dilution, lost-in-the-middle effects, and incorrect row--column binding~\citep{wang2025needleinatable}. As a result, models often fail to identify the correct evidence or produce inconsistent intermediate reasoning steps, even when the necessary information is present~\citep{li-etal-2025-longtablebench, wu2025tablebench}.


Existing approaches to LLM-based table reasoning largely follow two paradigms. The first treats reasoning as executable program generation, most commonly through Text-to-SQL translation~\citep{yu-etal-2018-spider,10.5555/3666122.3667957, cheng2023binder}. By delegating computation to database engines, these systems achieve strong performance on structured and aggregation-heavy queries (e.g.,  counting, sorting, and numerical comparison)~\citep{nahid-rafiei-2024-normtab}. However, symbolic execution relies on constructing precise logical predicates and therefore struggles when tables contain noisy textual descriptions, free-form attributes, or semi-structured content commonly observed in real-world data~\citep{abhyankar-etal-2025-h}.

The second paradigm performs direct reasoning by prompting LLMs with serialized tables~\citep{herzig2020tapas, wang2024chainoftable}. While this approach offers flexibility under ambiguity and noise, its performance degrades as table size increases, making it difficult for LLMs to capture row--column relationships and identify the correct cell values~\citep{10.1145/3539618.3591708, chen-2023-large}.

Retrieval and pruning strategies partially mitigate this issue by selecting subsets of rows or columns~\citep{abhyankar-etal-2025-h,wang2025tabsd}, but they still require reasoning within a single monolithic view and remain vulnerable to incomplete or fragmented evidence selection~\cite{lin-etal-2023-inner, nahid-rafiei-2024-tabsqlify}.

\begin{figure}[h]
\centering
\resizebox{0.95\linewidth}{!}{
\includegraphics[width=1\textwidth]{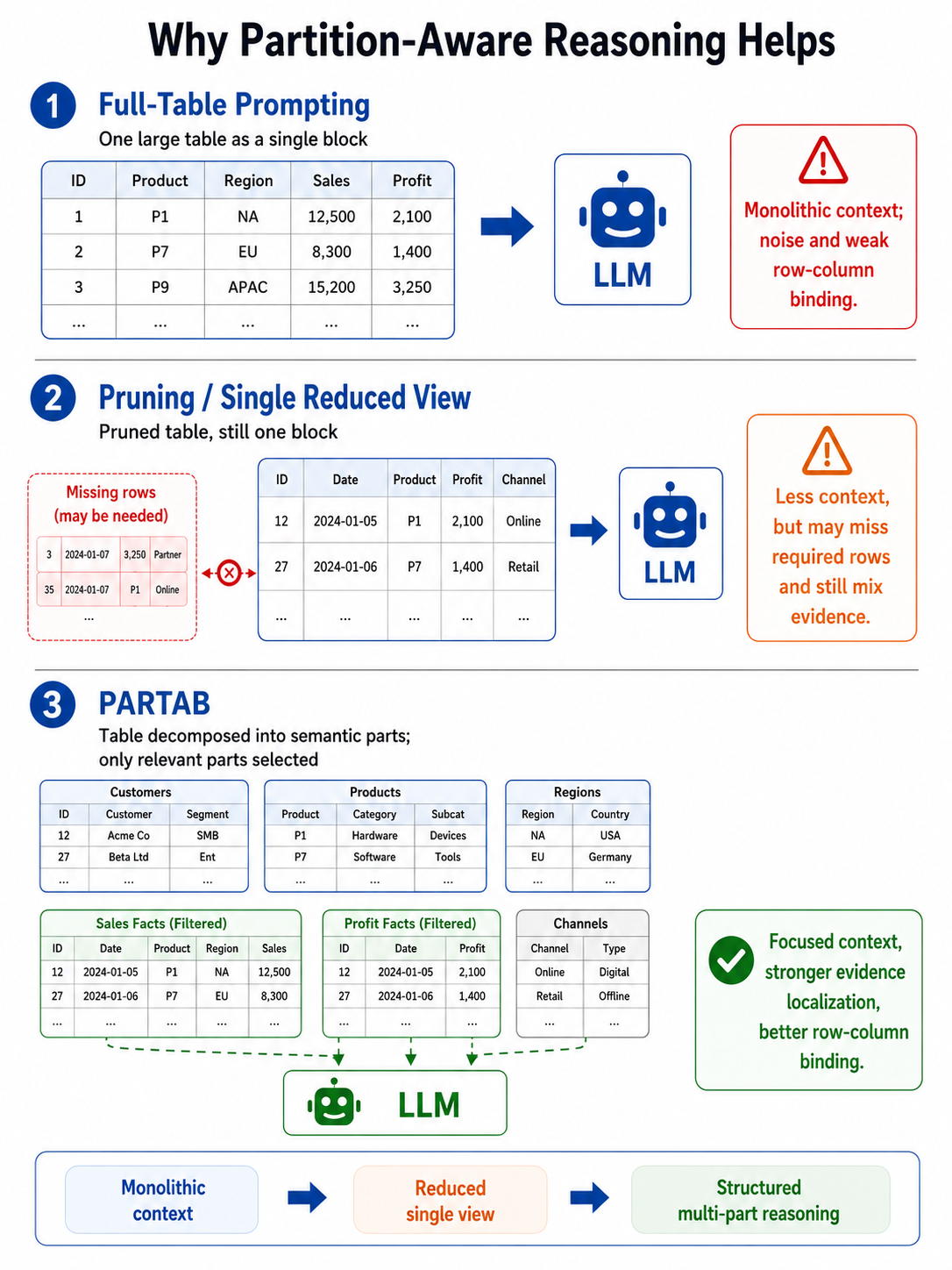}
}
\vspace{-1.0em}
\caption{Comparison of reasoning behaviors under different table reasoning paradigms.
\textbf{(I) Full-table prompting} exposes the model to the entire table, increasing irrelevant context and making evidence localization more difficult in long and wide tables. \textbf{(II) Single-view pruning} reduces context size but may remove required evidence, leading to incomplete reasoning and incorrect aggregation. \textit{All data shown are synthetic and for illustration only.}}
\vspace{-1.0em}
\label{fig:partab-comparison}
\end{figure}





\begin{figure*}[t]
\centering
\includegraphics[width=0.90\textwidth]{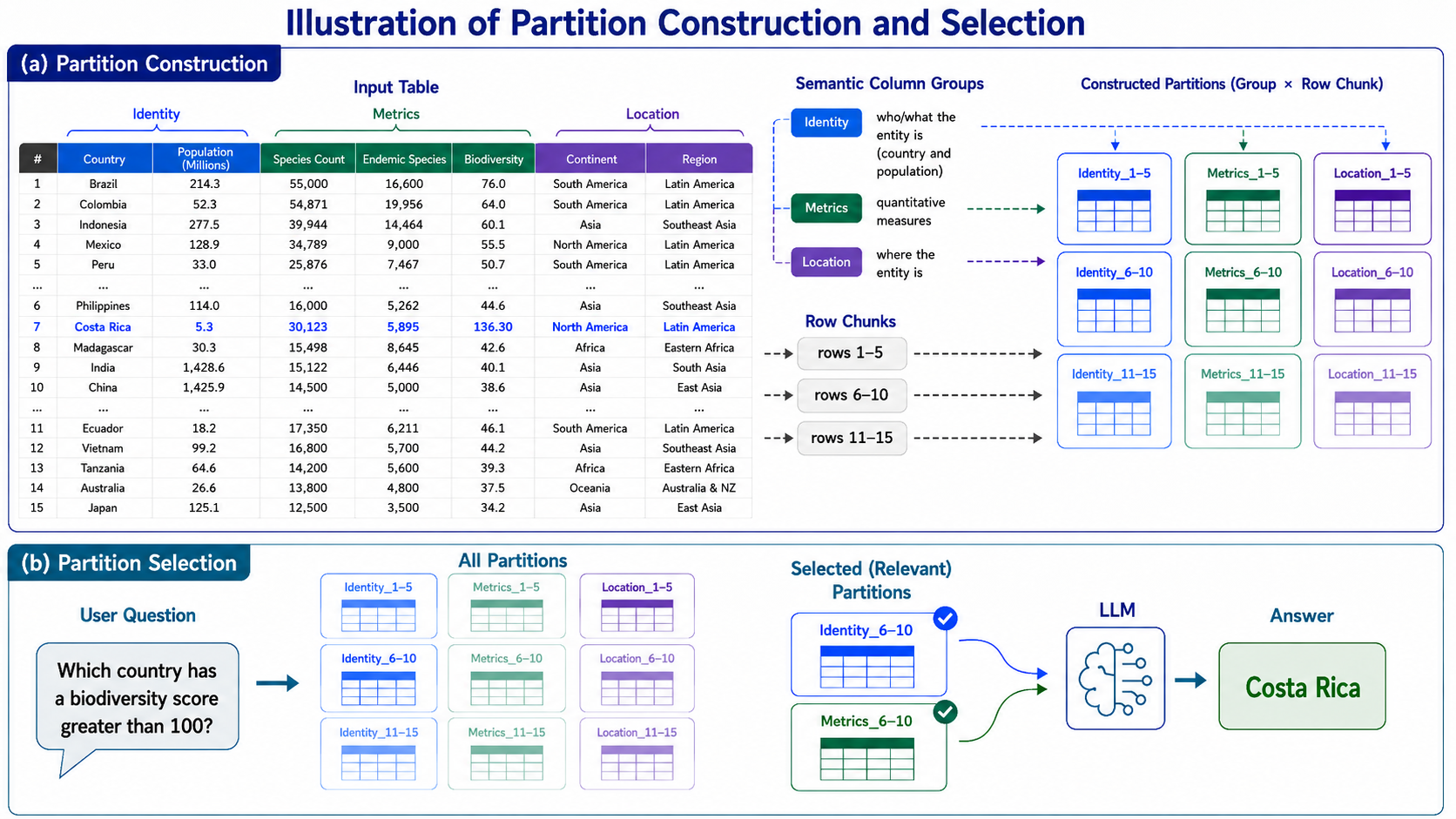}
\caption{Overview of PARTAB. Given a question and table, PARTAB constructs semantically coherent table partitions and performs localized understanding over selected partitions enabling reliable reasoning over large and noisy tables. \textit{All data shown are synthetic and for illustration only}.  
}
\vspace{-1.0em}
\label{fig:partab-method}
\end{figure*}

These limitations suggest that the core issue is not only \emph{what} information is retrieved, but \emph{how} reasoning is structured. 
We argue that table reasoning should be reframed as a \textbf{partitioned reasoning problem}, where reasoning is performed over semantically coherent subsets of the table rather than a single global view~(see Figure~\ref{fig:partab-comparison}). This perspective leads to three key research questions: \circnum{\textbf{1}} how can we provide LLMs with only the relevant evidences (\textbf{table parts}) of a table for a given query? \circnum{\textbf{2}} how can we construct partitions that preserve semantic coherence and relational structure? and \circnum{\textbf{3}} how can we select the minimal set of partitions that jointly contain sufficient evidence for correct reasoning?

To address these questions, we introduce \textbf{PARTAB} \emph{(\textbf{P}artition-\textbf{A}ware \textbf{R}easoning over \textbf{Tab}les)}, a framework that introduces a structured evidence interface between the LLM and the table. Rather than collapsing selected evidence into a single reduced table, PARTAB constructs a query-conditioned evidence state consisting of independently addressable, row-linked table regions. The LLM then reasons over this structured state by composing evidence across the selected regions. In this way, partitioning serves not merely as a preprocessing operation for context reduction, but as an explicit mechanism for organizing and controlling table reasoning.

We evaluate PARTAB on multiple challenging table reasoning benchmarks, including question answering and fact verification tasks. Results demonstrate consistent gains over full-table prompting and pruning-based baselines, outperforming several recent methods, particularly on large and noisy tables. Further analysis demonstrate that partition-aware reasoning improves evidence localization and yields more interpretable intermediate reasoning traces. 


Our contributions are threefold. \circnum{\textbf{1}} We introduce \textbf{PARTAB}, which constructs a query-conditioned structured evidence state consisting of semantically organized, independently addressable, and row-linked table regions. \circnum{\textbf{2}} We introduce a modular pipeline that combines semantic grouping, group selection, and row-linked part selection to improve alignment and reduce noise during reasoning. \circnum{\textbf{3}} We demonstrate that the benefits of partition-aware reasoning increase with table size and structural complexity, supporting its use as a scalable reasoning interface for large and noisy tables.






\section{Methodology}


We implement \textbf{PARTAB} as a modular partition-aware reasoning pipeline for table question answering (see \textbf{Figure~\ref{fig:partab-method}}). Given a question $q$ and a table $T$, the system produces an answer $a$ through four stages:
(1) \textit{Question Analyzer},
(2) \textit{Partition Builder},
(3) \textit{Group and Part Selector},
(4) \textit{Answer Executor}.

The key idea is to replace single-view reasoning over the entire table with \textbf{localized reasoning over structured table parts}. For example, a question about ``the highest population city'' should only involve the columns related to city and population, and a small subset of relevant rows, rather than the full table. \textbf{Algorithm~\ref{alg:partab}} summarizes the  pipeline. 


\begin{algorithm}[t]
\caption{\textsc{PARTAB}: Partition-Aware Table Reasoning}
\label{alg:partab}
\small
\begin{algorithmic}[1]
\Require Question $q$ and table $T$
\Ensure Predicted answer $a$

\State $z_q \gets \Call{Analyze}{q}$
\State $\Pi(T) \gets \Call{BuildPartitions}{T,q}$

\State $G_q \gets \Call{SelectGroups}{q,z_q,\Pi(T)}$
\State $\mathcal{P}_q \gets \Call{SelectParts}{q,G_q,\Pi(T)}$

\If{$\mathcal{P}_q = Null $}
    \State $\mathcal{P}_q \gets \Call{BasicSelect}{G_q,\Pi(T)}$
\EndIf

\State $a \gets \Call{Execute}{q,\mathcal{P}_q}$
\State \Return $a$
\end{algorithmic}
\end{algorithm}

\subsection{Table Preprocessing}

We first normalize the input table to ensure consistent schema representation. Column names are cleaned by lowercasing, removing or replacing special characters (e.g., replacing whitespace with underscores), and standardizing formatting. We then insert a \texttt{row\_id} column when absent:
\[
T' = \texttt{EnsureRowID}(\texttt{CleanColumns}(T)).
\]

The \texttt{row\_id} serves as a stable key for linking information across partitions in later stages.

\subsection{Question Analyzer}
In this stage, we aim to identify the reasoning structure of the question to guide downstream partitioning and selection.
Given a question $q$, an LLM-based analyzer produces a structured representation:
\[
z_q = \texttt{Analyze}(q),
\]
which encodes attributes such as question type (lookup, aggregation, comparison, etc.), required data processing operations (filtering, sorting, aggregation), and expected answer type. 

The intuition is that different questions require different table regions.
For example, an aggregation query focuses on filtered rows that should not be part of the aggregation, while a comparison query requires multiple candidate rows. The analyzer provides lightweight signals that guide both partition selection and reasoning.

\subsection{Partition Builder}
This stage decomposes the table into semantically coherent parts to enable localized reasoning. Rather than operating on a single serialized table, PARTAB constructs multiple structured views by partitioning both columns and rows.

\paragraph{Semantic Column Grouping.}
Columns are grouped into semantically coherent sets:
\[
G = \{g_1, g_2, \dots, g_m\}.
\]
Each non-key column must belong to exactly one group, while \texttt{row\_id} acts as a universal key. Every column that is not assigned by the mode is placed into a fallback \texttt{other} group. For example, in a table of countries, columns such as \texttt{population}, \texttt{area}, and \texttt{gdp} may form a ``statistics'' group, while \texttt{name} and \texttt{region} form a ``metadata'' group.

\paragraph{Chunked Part Construction.}
Each semantic group $g_i$ is further partitioned along the row dimension into fixed-size chunks of size $c$, where each chunk is identified by the corresponding range of \texttt{row\_id}s. Formally, 
%
\[
P_{i,j} = T'[r_{j:j+c-1}, \{\texttt{row\_id}\} \cup g_i],
\]
where $i$ indexes the semantic group and $j$ indexes the row chunk. Each part is stored together with its associated metadata, including the group name, row range, column set, and the corresponding data content. In our implementation, we use a default chunk size of $c=5$. This stage creates multiple localized views of the table. Unlike pruning, which produces a single reduced table, PARTAB produces a \emph{set of complementary parts}, enabling flexible evidence composition.

\subsection{Group and Part Selector}
This stage identifies the minimal set of table regions required for answering the question.

\paragraph{Group Selection.}
We first select relevant column groups:
\[
G_q \subseteq G.
\]

The selector is implemented as an LLM module that receives:
(1) the question,
(2) the question analysis output, and
(3) the list of available column groups.
It is instructed to select only the groups necessary for answering the question. This step reduces the search space before fine-grained part selection and acts as a schema-level routing mechanism. For example, a population-related query should ignore groups such as timestamps or descriptions.

\paragraph{Part Selection.}
We then select specific row chunks within the chosen groups:
\[
\Pi_q(T) = \{P \in \Pi(T) \mid \texttt{group}(P) \in G_q\}.
\]

We consider three selection strategies:

\emph{Basic Selection.}
This mode selects all candidate parts:
\[
\mathcal{P}_q = \Pi_q(T),
\]
and it serves as the simplest partition-aware baseline.

\emph{Similarity-Based Selection.}
This strategy ranks parts using TF--IDF similarity with the question and select top-$k$:
\[
\mathcal{P}_q = \texttt{TopK}_{\text{TF-IDF}}(q, \Pi_q(T)).
\]

\emph{LLM-Based Selection.}
In this mode, an LLM is used to directly predict the minimal set of relevant parts. 

This stage ensures that reasoning operates over a \emph{focused yet sufficient} subset of the table, reducing noise while preserving necessary evidence.

\subsection{Answer Executor}

This stage performs reasoning over selected parts to produce the final answer.
Here selected parts are serialized into a compact context:
\[
C_q = \texttt{Serialize}(\mathcal{P}_q).
\]
\[
a = \texttt{Execute}(q, C_q).
\]

Each part is rendered with metadata (group, row range, columns) and a preview of rows.
The prompt explicitly instructs the model to answer using only the provided table parts, treat \texttt{row\_id} as the linking key across parts, avoid relying on outside knowledge, and return only a short final answer. If the answer cannot be determined from the selected parts, the model is instructed to return a fixed abstention response.






\begin{table*}[t]
\centering
\small
\setlength{\tabcolsep}{8pt}
\begin{tabular}{l|c|c|cc}
\toprule

\multirow{2}{*}{Method} 
& WikiTableQuestions 
& TabFact 
& \multicolumn{2}{c}{TableBench} \\
& EM & Acc & NR (EM) & FC (EM) \\

\midrule
BINDER \citep{cheng2023binder}	                         & 38.03 & 67.50 &- & - \\
DATER \citep{10.1145/3539618.3591708}	                          & 52.00  & 74.70 &- & - \\
ReAcTable \citep{10.14778/3659437.3659452}	                      & 57.09 & - &- & - \\ 
\midrule

End-to-End Prompting                    & 59.43 & 73.22 & 59.50 & 66.66 \\
Chain-of-Thought (CoT)                 & 64.31 & 75.99 & - & - \\

Chain-of-Table~\cite{wang2024chainoftable} & 68.53 & 85.09 & - & - \\
TabSQLify~\cite{nahid-rafiei-2024-tabsqlify}    & 68.74 & 78.30 & 66.25 & 70.83 \\
Rank\_Agent  \citep{wu2025tablebench}      & - & - & \textbf{78.34} & \textbf{87.50} \\
PoTable \cite{mao2026potablesystematicthinkingplanthenexecute}  & {64.73} & {88.93} & - & - \\
H-Star \citep{abhyankar-etal-2025-h}   & {74.93} & {89.42} & 69.52 & 79.17 \\
TableMaster \citep{cao2026tablemasterrecipeadvancetable}  & \underline{78.13} & \underline{90.12} & - & - \\
\midrule
\rowcolor{gray!10}
\textbf{PARTAB (Ours)}             & \textbf{79.31} & \textbf{90.48} & \underline{70.33} & \underline{82.71}  \\
\bottomrule
\end{tabular}
\caption{Downstream performance on table reasoning benchmarks. PARTAB achieves strong results across question answering, fact verification, and numerical reasoning, outperforming several recent baselines on WikiTableQuestions and TabFact while remaining competitive on TableBench.}
\vspace{-1em}
\label{tab:main_results}
\end{table*}


\section{Evaluation and Experimental Setup}

\paragraph{Datasets.}

We evaluate PARTAB on three benchmarks covering both table question answering and fact verification tasks. \textbf{WikiTableQuestions}~\citep{pasupat-liang-2015-compositional} evaluates compositional reasoning over Wikipedia tables, and we report {Exact Match (EM)} accuracy following standard protocols. 
\textbf{TabFact}~\citep{2019TabFactA} focuses on table-based fact verification, where the task is to determine whether a statement is entailed or refuted by a table; we report {classification accuracy (Acc)}. 
\textbf{TableBench}~\citep{wu2025tablebench} evaluates reasoning over large and complex tables and includes two subsets: \textbf{Numerical Reasoning (NR)} and \textbf{Fact Checking (FC)}. We report {Exact Match (EM)} for both NR and FC tasks.



\paragraph{Implementation.}

We adopt a modular design in PARTAB, where different components are instantiated with models suited to their functional roles. We use the lightweight and cost-efficient \texttt{GPT-5-nano} for partition construction to limit inference cost, and \texttt{GPT-4o-mini} for answer execution to enable fair comparison with prior work to demonstrate the effect of partition-aware reasoning. To assess robustness across model families, we evaluate our method using two further LLM backbones, \texttt{Gemini-2.5-Flash-Lite} and \texttt{DeepSeek-v4-Flash}. All answer-generation experiments use zero-shot prompting with concise instructions, without few-shot demonstrations, chain-of-thought, or program-of-thought prompting. Detailed prompts and illustrative examples are provided in \textbf{Appendix~\ref{app:example_prompts}}. 




The default configuration sets the row chunk size to $c=5$ and uses top-$k=6$ for similarity-based retrieval, where similarity is computed using a basic TF-IDF representation. \footnote{We will release the code and data upon acceptance of the paper.}


\section{Results and Discussion}

We analyze PARTAB through three empirical questions:
\circnum{\textbf{1}} whether gains come from structured evidence construction rather than additional inference budget;
\circnum{\textbf{2}} the contribution of semantic partitioning and hierarchical selection; and
\circnum{\textbf{3}} how the benefit of partition-aware reasoning changes with table complexity.

\paragraph{Downstream Results.}

Table~\ref{tab:main_results} shows that PARTAB performs strongly across diverse table reasoning tasks. It achieves the best result on WikiTableQuestions (79.31 EM), surpassing TableMaster (78.13) and earlier methods, and reaches 90.48 accuracy on TabFact, indicating effective evidence localization and semantic grounding. On TableBench, PARTAB remains competitive with 70.33 EM on numerical reasoning and 82.71 accuracy on fact checking. Although specialized systems such as Rank\_Agent perform better on aggregation-heavy tasks, PARTAB achieves strong results without full-table execution. Overall, PARTAB provides a balanced and robust approach across semantic and numerical reasoning by improving accuracy through localized reasoning over partitioned table regions.


\begin{figure}[h]
    \centering
    \resizebox{.85\linewidth}{!}{
    \includegraphics[width=\linewidth]{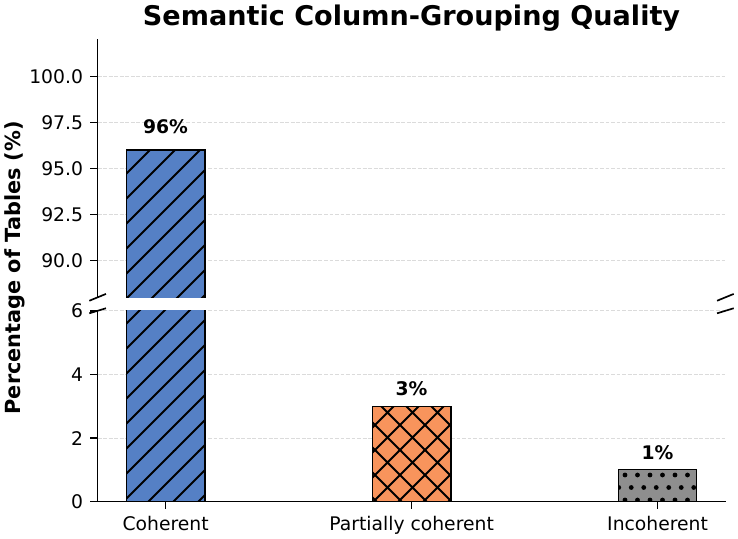}
    }
    \caption{Partition construction performance. Manual evaluation of semantic column-grouping quality over 100 tables.The generated groups are fully coherent in 96\% of the evaluated cases. }
    \vspace{-1.0em}
    \label{fig:partition_construction_quality}
\end{figure}


\paragraph{Partition Construction Performance.}

We evaluate the quality of partition construction by assessing the semantic coherence of the generated column groups. Since column grouping is produced by an LLM without a unique ground-truth decomposition, we conduct a manual evaluation to determine whether the induced groups align with meaningful semantic structures in the table. As shown in Figure~\ref{fig:partition_construction_quality}, 96\% of the groups are fully coherent, indicating that the model reliably captures high-level schema structure. The remaining 4\% of errors mainly involve semantically related columns being separated or unrelated columns being grouped together. Overall, these results show that PARTAB constructs robust and meaningful partitions for downstream reasoning.




\begin{table}[h]
\centering
\small
\setlength{\tabcolsep}{2pt}
\renewcommand{\arraystretch}{1.08}

\resizebox{\linewidth}{!}{
\begin{tabular}{l|c|c|c|c}
\toprule
{Selection Strategy}
& {WikiTQ}
& {TabFact}
& \multicolumn{2}{c}{{TableBench}} \\
\cmidrule(lr){4-5}
& {EM}
& {Acc}
& {NR (EM)}
& {FC (EM)} \\
\midrule
Basic Selection
& \underline{74.71}
& \underline{87.88}
& \underline{64.43}
& \underline{79.26} \\

Similarity-based Selection
& 67.25
& 85.87
& 58.69
& 59.25 \\

LLM-based Selection
& \textbf{79.31}
& \textbf{90.48}
& \textbf{70.33}
& \textbf{82.71} \\
\bottomrule
\end{tabular}
}
\caption{
Comparison of part-selection strategies across table reasoning benchmarks.
The best result in each column is shown in bold.
}
\vspace{-1.0em}
\label{tab:partition_selection}
\end{table}

\paragraph{Partition Selection Performance.}

Table~\ref{tab:partition_selection} shows that effective part selection is critical for partition-aware reasoning. LLM-based selection achieves the best performance across all benchmarks, including WikiTableQuestions (79.31 EM), TabFact (90.48 Acc), and TableBench (70.33 / 82.71), demonstrating the importance of semantic understanding in identifying minimal yet sufficient partitions.

In contrast, similarity-based selection performs the worst, indicating that surface-level similarity is insufficient for capturing true relevance, particularly for compositional queries. Basic selection performs better but still includes unnecessary context, which can reduce reasoning precision compared to more targeted selection.


\begin{figure}[t]
    \centering
    \resizebox{.87\linewidth}{!}{
    \includegraphics[width=0.85\textwidth]{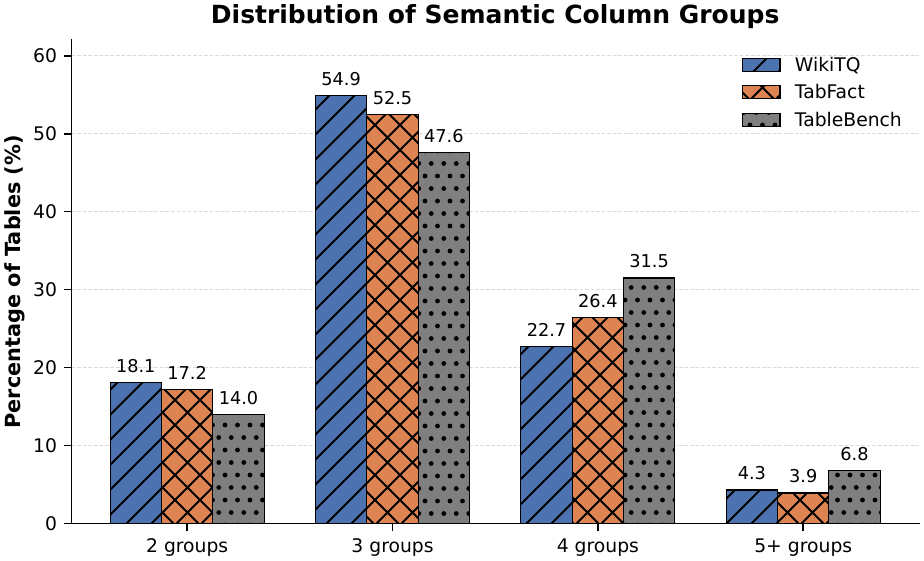}
    }
    \caption{
    Distribution of the number of semantic column groups generated per
    table. Most tables are decomposed into three or four groups across all
    evaluated datasets.
    }
    \vspace{-1.0em}
    \label{fig:group_distribution}
\end{figure}

\paragraph{Distribution of Semantic Column Groups.}
\label{app:group_distribution}

The Column Grouping prompt provides a soft preference for constructing between two and five semantic groups. This is not a hard constraint; the model may produce additional groups when required by a more complex schema.

Figure~\ref{fig:group_distribution} shows that most tables are represented by three or four semantic groups. For WikiTQ, 77.6\% of tables contain three or four groups; the corresponding proportions are 78.9\% for TabFact and 79.1\% for TableBench. Six groups are generated in at most 0.2\% of examples.

These results indicate that the grouping procedure produces compact and
interpretable schema organizations rather than excessively fragmenting thetable. At the same time, the model retains the ability to generate more fine-grained groupings for unusually wide or semantically diverse schemas.

\begin{table}[ht]
\centering
\small
\setlength{\tabcolsep}{4pt}
\resizebox{\linewidth}{!}{
\begin{tabular}{l|c|c|c|c}
\toprule
{LLM Backbone}
& {WikiTQ}
& {TabFact}
& \multicolumn{2}{c}{{TableBench}} \\
& EM & Acc & NR (EM) & FC (EM) \\
\midrule
\texttt{GPT-4o-mini}
& \textbf{79.31}
& \underline{90.48}
& \textbf{70.33}
& \textbf{82.71} \\

\texttt{Gemini-2.5-Flash-Lite}
& 71.23
& 83.56
& 61.29
& 64.53 \\

\texttt{DeepSeek-v4-Flash}
& \underline{78.30}
& \textbf{91.70}
& \underline{70.23}
& \underline{75.34} \\
\bottomrule
\end{tabular}}
\caption{
Performance of \textsc{PARTAB} across different answer-execution
backbones.}
\vspace{-1.0em}
\label{tab:llm_backbone_results}
\end{table}


\paragraph{Robustness Across LLM Backbones.}
\label{app:backbone_robustness}

We evaluate \textsc{PARTAB} using three answer-execution backbones with
different capabilities and model families. Table~\ref{tab:llm_backbone_results}
shows that the framework remains effective across all three models.
DeepSeek-v4-Flash closely matches GPT-4o-mini on WikiTQ and
TableBench-NR and obtains the highest TabFact accuracy. Although
Gemini-2.5-Flash-Lite produces lower absolute performance, it remains capable
of following the same partition-aware reasoning pipeline.

These results indicate that \textsc{PARTAB} is not tied to a single
proprietary model. Nevertheless, the performance variation across backbones
shows that final accuracy still depends on the executor's ability to follow
structured instructions, maintain row--column alignment, perform numerical
operations, and reason faithfully over selected table parts.

\begin{table*}[ht]
\centering
\small
\setlength{\tabcolsep}{8pt}

\resizebox{0.75\linewidth}{!}{
\begin{tabular}{l|l|cc|cc|cc}
\toprule
\multirow{2}{*}{{LLM Backbone}}
& \multirow{2}{*}{{Method}}
& \multicolumn{2}{c|}{{WikiTQ-Hard}}
& \multicolumn{2}{c|}{{TabFact-Hard}}
& \multicolumn{2}{c}{{TB-NR-Hard}} \\
\cmidrule(lr){3-4}
\cmidrule(lr){5-6}
\cmidrule(lr){7-8}
&
& {EM}
& ${\Delta}$
& {EM}
& ${\Delta}$
& {EM}
& ${\Delta}$ \\
\midrule

\multirow{2}{*}{\texttt{GPT-4o-mini}}
& Full Table
& 45.80
& --
& 55.32
& --
& 40.47
& -- \\
& {{PARTAB} (Ours)}
& \textbf{55.72}
& \textbf{+9.92}
& \textbf{80.85}
& \textbf{+25.53}
& \textbf{61.90}
& \textbf{+21.43} \\
\midrule

\multirow{2}{*}{\texttt{GPT-5-mini}}
& Full Table
& 60.30
& --
& 57.14
& --
& 57.45
& -- \\
& {{PARTAB} (Ours)}
& \textbf{61.83}
& \textbf{+1.53}
& \textbf{78.26}
& \textbf{+21.12}
& \textbf{61.90}
& \textbf{+4.45} \\
\midrule

\multirow{2}{*}{\texttt{DeepSeek-v4-Pro}}
& Full Table
& 58.01
& --
& 55.32
& --
& 54.76
& -- \\
& {{PARTAB} (Ours)}
& \textbf{59.54}
& \textbf{+1.53}
& \textbf{89.36}
& \textbf{+34.04}
& \textbf{57.14}
& \textbf{+2.38} \\

\bottomrule
\end{tabular}
}

\vspace{-0.2em}
\caption{
Performance of full-table prompting and \textsc{PARTAB} on hard subsets
containing large or wide tables. WikiTQ-Hard and TB-NR-Hard are evaluated
using Exact Match (EM), while TabFact-Hard is evaluated using accuracy.
Each $\Delta$ denotes the absolute improvement of \textsc{PARTAB} over
full-table prompting with the same backbone.
}
\vspace{-1.0em}
\label{tab:hard_subset_frontier}
\end{table*}


\paragraph{Frontier Models on Large and Complex Tables.}
\label{app:hard_subset}

To examine whether partition-aware reasoning remains useful when the executor
has stronger reasoning capabilities and a larger context window, we construct
a hard subset for each benchmark. An example is included in the hard subset
when its table contains more than 45 rows or more than 10 columns. These examples are particularly challenging due to attention dilution, increased irrelevant context, and greater difficulty in evidence localization. 

Table~\ref{tab:hard_subset_frontier} compares \textsc{PARTAB} with full-table
prompting using the same backbone. \textsc{PARTAB} improves performance for
every backbone--dataset combination. The average improvement is
$+18.96$ points for GPT-4o-mini, $+9.03$ points for GPT-5-mini, and
$+12.65$ points for DeepSeek-v4-Pro.

The largest improvements are observed on TabFact-Hard, where
\textsc{PARTAB} improves accuracy by $25.53$, $21.12$, and $34.04$ points
for GPT-4o-mini, GPT-5-mini, and DeepSeek-v4-Pro, respectively. The gains on
WikiTQ-Hard and TableBench-NR-Hard become smaller for stronger models,
suggesting that frontier models can partially compensate for long-context
difficulty on some compositional and numerical questions. However, the consistent improvements show that stronger reasoning capabilities and larger context windows do not fully eliminate challenges in evidence localization and reasoning over large tables.


\paragraph{Partition Structure and Context Reduction.}
\label{app:partition_statistics}

We analyze the structures generated by the Partition Builder across the
three evaluation datasets. Table~\ref{tab:partition_statistics} reports the
average number of semantic column groups, columns per group, candidate
partitions, and selected partitions.

Across datasets, each table is decomposed into approximately 3.1--3.3
semantic groups, with roughly two non-key columns in each group. Although
partition construction creates between 9.84 and 17.56 candidate parts per
table, the selector retains only 2.44--3.88 parts on average.

Relative to the candidate partition set, this corresponds to a reduction of
77.9\% on WikiTQ, 75.2\% on TabFact, and 75.2\% on TableBench. Therefore,
\textsc{PARTAB} exposes the final answer executor to only a small fraction
of the available table regions.

\begin{table}[ht]
\centering
\small
\setlength{\tabcolsep}{4.5pt}
\renewcommand{\arraystretch}{1.08}

\resizebox{\linewidth}{!}{
\begin{tabular}{l|ccc|cc}
\toprule
\multirow{2}{*}{{Dataset}}
& \multicolumn{3}{c}{{Partition Construction}}
& \multicolumn{2}{c}{{Partition Selection}} \\
\cmidrule(lr){2-4}
\cmidrule(lr){5-6}
& \shortstack{{Groups}\\{/ Table}}
& \shortstack{{Columns}\\{/ Group}}
& \shortstack{{Parts}\\{/ Table}}
& \shortstack{{Selected}\\{Parts}}
& \shortstack{{Reduction}\\{(\%)}} \\
\midrule
WikiTQ
& 3.13
& 2.10
& 17.56
& 3.88
& 77.9 \\

TabFact
& 3.17
& 2.11
& 9.84
& 2.44
& 75.2 \\

TableBench
& 3.32
& 2.31
& 12.27
& 3.04
& 75.2 \\
\bottomrule
\end{tabular}
}

\caption{
Statistics of partition construction and selection.
Reduction is computed as
$1 - \frac{\text{selected parts}}{\text{candidate parts}}$.
} 
\vspace{-1.0em}
\label{tab:partition_statistics}
\end{table}


\paragraph{Component-Level Ablations.}
\label{app:component_ablations}

We conduct component-level ablations on WikiTQ to isolate the contribution
of the Question Analyzer, semantic column grouping, and joint row--column
partitioning. The results are shown in
Table~\ref{tab:component_ablations}.

\begin{table}[ht]
\centering
\small
\setlength{\tabcolsep}{6pt}
\renewcommand{\arraystretch}{1.08}

\resizebox{\linewidth}{!}{
\begin{tabular}{l|c|c}
\toprule
\multirow{2}{*}{{Configuration}}
& \multicolumn{2}{c}{{WikiTQ}} \\
\cmidrule(lr){2-3}
& {EM}
& ${\Delta}$ \\
\midrule
Full \textsc{PARTAB}
& \textbf{79.31}
& -- \\

\midrule

\textsc{PARTAB} w/o Question Analyzer
& 72.21
& $-7.10$ \\

\textsc{PARTAB} with Random Column Grouping
& 68.84
& $-10.47$ \\

\textsc{PARTAB} with Row Chunking Only
& 57.19
& $-22.12$ \\
\bottomrule
\end{tabular}
}

\caption{
Component-level ablation results on WikiTQ. Each $\Delta$ denotes the
absolute change in exact match relative to the full \textsc{PARTAB}
configuration.
}
\vspace{-1.0em}
\label{tab:component_ablations}
\end{table}

Removing the Question Analyzer decreases exact match by $7.10$ points,
showing that the structured question representation improves schema-level
routing and partition selection. Replacing semantic column groups with
randomly constructed groups reduces performance by $10.47$ points, directly
demonstrating that semantic coherence is important rather than grouping
alone.

\begin{figure}[h]
    \centering
    \resizebox{0.8\linewidth}{!}{
    \includegraphics[width=\linewidth]{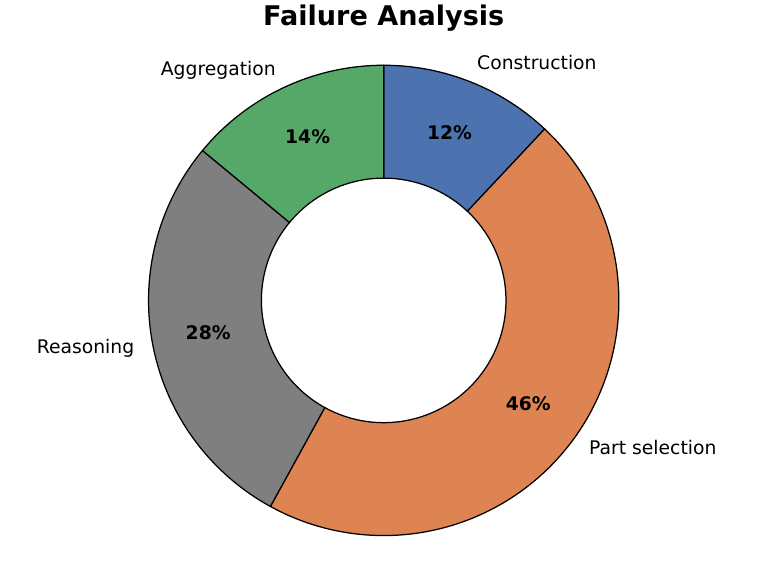}
    }
    \caption{
    Distribution of failure types based on 100 incorrect
    predictions. Part-selection errors constitute the largest source
    of failure.}
    \vspace{-1.0em}
    \label{fig:error_analysis}
\end{figure}

The largest degradation occurs when \textsc{PARTAB} uses row chunking
without semantic column grouping, resulting in a $22.12$-point reduction.
This result indicates that row segmentation alone does not provide the
structured evidence organization required for reliable table reasoning.
Instead, the performance gains arise from jointly localizing relevant rows
and semantically related columns.

\paragraph{Error Analysis.}
\label{app:error_analysis}

We manually analyze 100 incorrect predictions and categorize the failures into four types (Figure~\ref{fig:error_analysis}). Part-selection errors are the most common (46\%), occurring when relevant partitions are missed or row coverage is insufficient, indicating that evidence localization remains the main bottleneck. Reasoning errors account for 28\%, where sufficient evidence is present but the executor fails in comparison, inference, or computation. Aggregation errors represent 14\% and mainly arise from questions requiring near-global row coverage. Construction errors are least frequent (12\%), consistent with the high semantic coherence of the generated column groups and suggesting that partition construction is more reliable than downstream selection. We provide additional experimental results, and analysis in the \textbf{Appendix~\ref{app:additional_results}} and \textbf{\ref{app:additional_analysis}}.

\section{Related Work}

\paragraph{LLM-based Table Reasoning.}
Large language models (LLMs) have enabled strong progress in table question answering and fact verification~\citep{cao2026tablemasterrecipeadvancetable, zhou2025tqasurvey,lu2025tablellmsurvey,fang2024tabularllm}. Early models such as TAPAS~\citep{herzig2020tapas} and TAPEX~\citep{liu2022tapex} demonstrated the effectiveness of pretraining for tabular reasoning, while recent prompting-based approaches improve flexibility but struggle with long tables due to attention limitations and incomplete evidence utilization~\citep{tyagi2024agentictableqa}. 

\paragraph{Text-to-SQL and Symbolic Reasoning.} Another line of work formulates table reasoning as Text-to-SQL generation~\citep{yu-etal-2018-spider,10.5555/3666122.3667957}, enabling reliable aggregation through executable queries. While effective for structured data, these approaches rely on precise predicate construction and often fail under noisy textual conditions~\citep{wang2024tabaf,evkarpidi2025bridging}.

\paragraph{Decomposition and Retrieval.}
To address scalability, prior work decomposes tables or retrieves relevant substructures. Methods such as TabSQLify~\citep{nahid-rafiei-2024-tabsqlify}, TabSD~\citep{wang2025tabsd}, and Tree-of-Table~\citep{ji2025treeoftable} reduce reasoning complexity through structured decomposition. Recent approaches further explore sub-table retrieval and structured reasoning, including ITR~\citep{lin-etal-2023-inner}, PieTa~\citep{lee2024piece}, RoT~\citep{zhang-etal-2025-rot}, and CABINET~\citep{patnaik2024cabinet}. Long-context benchmarks highlight persistent performance degradation as table size increases~\citep{li-etal-2025-longtablebench,wang2025needleinatable, wu2025tablebench}.
Recent retrieval-augmented approaches such as TableRAG~\citep{chen2024tablerag,yu-etal-2025-tablerag} improve scalability by retrieving relevant table content, but do not explicitly structure reasoning within tables.

\paragraph{Hybrid and Tool-Augmented Reasoning.}
Hybrid frameworks combine LLM reasoning with external tools such as SQL or programs. Representative approaches include ReAcTable~\citep{10.14778/3659437.3659452}, Weaver~\citep{khoja-etal-2025-weaver}, E$^5$~\citep{zhang-etal-2024-e5}, TIDE~\citep{yang2025triples}, and TABDSR~\citep{jiang-etal-2025-tabdsr}, as well as broader neuro-symbolic systems~\citep{zhang-etal-2024-syntqa,cheng2023binder,zhang2025tablemoeneurosymbolicroutingstructured,10.1145/3539618.3591708,abhyankar-etal-2025-h}. These methods improve numerical reasoning and interpretability but often depend on execution pipelines or structured representations.


Existing approaches typically rely on full-table reasoning, single-view decomposition or retrieval, or symbolic execution, each involving trade-offs among flexibility, scalability, and evidence completeness. \textbf{PARTAB} instead constructs a structured evidence interface in which query-relevant information is organized into semantically coherent, row-linked table regions. By combining semantic grouping with hierarchical partition selection and cross-region evidence composition, PARTAB enables localized reasoning while preserving the relational structure needed to connect evidence across the table.



\section{Conclusion}

We introduced \textbf{PARTAB}, a partition-aware framework that constructs a structured evidence interface between LLMs and tables. PARTAB organizes query-relevant evidence into semantically coherent, row-linked regions and composes the selected regions for reasoning rather than relying on a full table or single reduced view. Experiments across multiple table reasoning benchmarks show strong performance, improved evidence localization, and substantial context reduction, with larger benefits on complex tables. These results demonstrate that structured, partition-aware evidence construction provides an effective reasoning interface for scalable table understanding.



\section*{Limitations}

PARTAB focuses on inference-time reasoning over tables, with particular emphasis on settings where the evidence required for a query is localized to a subset of rows and columns. The framework is designed to improve evidence localization and structured reasoning over large and noisy tables, rather than to replace symbolic execution for exhaustive aggregation or to address broader settings such as complex multi-table operations, temporal reasoning, or database-scale query processing.

Despite its effectiveness, PARTAB has several limitations. First, it relies on LLM-based components for question analysis, grouping, and selection, making it sensitive to prompt design and model variability, with errors potentially propagating across stages. Second, PARTAB does not explicitly enforce global completeness for aggregation tasks, which can lead to missing evidence when full-table coverage is required. Third, the current heuristic partitioning strategy (fixed chunk size and LLM-based grouping) may not generalize optimally across diverse table structures. Finally, although each step operates on reduced context, the multi-stage pipeline can introduce additional latency compared to single-pass prompting.

Future work includes integrating symbolic execution for reliable aggregation, improving part selection through learned or hybrid retrieval mechanisms, and extending PARTAB to more complex settings such as multi-table reasoning and temporal queries.

\section*{Ethics Statement}

PARTAB is a general-purpose framework for table reasoning and does not
introduce or collect sensitive user data. However, like other LLM-based
reasoning systems, it may produce incorrect answers when relevant evidence
is omitted or downstream reasoning fails. Such errors could be consequential
if the system were applied without verification in high-stakes domains.
PARTAB also requires multiple LLM inference calls, introducing additional
computational cost relative to single-pass prompting, although our experiments
show substantially lower token usage than several multi-stage baselines.
We therefore view PARTAB primarily as a research framework and recommend
appropriate validation and human oversight for consequential applications.



\bibliography{custom}

\begin{thebibliography}{41}
\providecommand{\natexlab}[1]{#1}

\bibitem[{Abhyankar et~al.(2025)Abhyankar, Gupta, Roth, and Reddy}]{abhyankar-etal-2025-h}
Nikhil Abhyankar, Vivek Gupta, Dan Roth, and Chandan~K. Reddy. 2025.
\newblock \href {https://doi.org/10.18653/v1/2025.naacl-long.445} {{H}-{STAR}: {LLM}-driven hybrid {SQL}-text adaptive reasoning on tables}.
\newblock In \emph{Proceedings of the 2025 Conference of the Nations of the Americas Chapter of the Association for Computational Linguistics: Human Language Technologies (Volume 1: Long Papers)}, pages 8841--8863, Albuquerque, New Mexico. Association for Computational Linguistics.

\bibitem[{Cao and Liu(2026)}]{cao2026tablemasterrecipeadvancetable}
Lang Cao and Hanbing Liu. 2026.
\newblock \href {https://arxiv.org/abs/2501.19378} {Tablemaster: A recipe to advance table understanding with language models}.
\newblock \emph{Preprint}, arXiv:2501.19378.

\bibitem[{Chen et~al.(2024)Chen, Miculicich, Eisenschlos, Wang, Wang, Chen, Fujii, Lin, Lee, and Pfister}]{chen2024tablerag}
Si-An Chen, Lesly Miculicich, Julian~Martin Eisenschlos, Zifeng Wang, Zilong Wang, Yanfei Chen, Yasuhisa Fujii, Hsuan-Tien Lin, Chen-Yu Lee, and Tomas Pfister. 2024.
\newblock \href {https://openreview.net/forum?id=41lovPOCo5} {Table{RAG}: Million-token table understanding with language models}.
\newblock In \emph{The Thirty-eighth Annual Conference on Neural Information Processing Systems}.

\bibitem[{Chen(2023)}]{chen-2023-large}
Wenhu Chen. 2023.
\newblock \href {https://doi.org/10.18653/v1/2023.findings-eacl.83} {Large language models are few(1)-shot table reasoners}.
\newblock In \emph{Findings of the Association for Computational Linguistics: EACL 2023}, pages 1120--1130, Dubrovnik, Croatia. Association for Computational Linguistics.

\bibitem[{Chen et~al.(2020)Chen, Wang, Chen, Zhang, Wang, Li, Zhou, and Wang}]{2019TabFactA}
Wenhu Chen, Hongmin Wang, Jianshu Chen, Yunkai Zhang, Hong Wang, Shiyang Li, Xiyou Zhou, and William~Yang Wang. 2020.
\newblock Tabfact : A large-scale dataset for table-based fact verification.
\newblock In \emph{International Conference on Learning Representations (ICLR)}, Addis Ababa, Ethiopia.

\bibitem[{Cheng et~al.(2023)Cheng, Jiang, Zhang, and Li}]{cheng2023binder}
Zhoujun Cheng, Tianyang Jiang, Qianhui Zhang, and Lei Li. 2023.
\newblock Binding language models in symbolic languages for table question answering.
\newblock In \emph{International Conference on Learning Representations (ICLR)}.

\bibitem[{Evkarpidi and Tutubalina(2025)}]{evkarpidi2025bridging}
Nikolas Evkarpidi and Elena Tutubalina. 2025.
\newblock \href {https://arxiv.org/abs/2506.09657} {Bridging the gap between open-source and proprietary large language models in table question answering}.

\bibitem[{Fang et~al.(2024)Fang, Xu, Tan, Zhang, Hu, Qi, Nickleach, Socolinsky, Sengamedu, and Faloutsos}]{fang2024tabularllm}
Xi~Fang, Weijie Xu, Fiona~Anting Tan, Jiani Zhang, Ziqing Hu, Yanjun Qi, Scott Nickleach, Diego Socolinsky, Srinivasan Sengamedu, and Christos Faloutsos. 2024.
\newblock Large language models on tabular data: Prediction, generation, and understanding.
\newblock \emph{ACM Computing Surveys}.

\bibitem[{Herzig et~al.(2020)Herzig, Nowak, Müller, Piccinno, and Eisenschlos}]{herzig2020tapas}
Jonathan Herzig, Pawel~Krzysztof Nowak, Thomas Müller, Francesco Piccinno, and Julian Eisenschlos. 2020.
\newblock \href {https://doi.org/10.18653/v1/2020.acl-main.398} {Tapas: Weakly supervised table parsing via pre-training}.
\newblock In \emph{Proceedings of the 58th Annual Meeting of the Association for Computational Linguistics}, page 4320–4333. Association for Computational Linguistics.

\bibitem[{Ji et~al.(2025)Ji, Zhu, Gao, Xu, Lu, Ye, and Zhao}]{ji2025treeoftable}
Deyi Ji, Lanyun Zhu, Siqi Gao, Peng Xu, Hongtao Lu, Jieping Ye, and Feng Zhao. 2025.
\newblock Tree-of-table: Unleashing the power of llms for enhanced large-scale table understanding.
\newblock In \emph{International Conference on Learning Representations (ICLR)}.

\bibitem[{Jiang et~al.(2025)Jiang, Yu, Chen, Lu, and Zeng}]{jiang-etal-2025-tabdsr}
Changjiang Jiang, Fengchang Yu, Haihua Chen, Wei Lu, and Jin Zeng. 2025.
\newblock \href {https://doi.org/10.18653/v1/2025.findings-emnlp.169} {{T}ab{DSR}: Decompose, sanitize, and reason for complex numerical reasoning in tabular data}.
\newblock In \emph{Findings of the Association for Computational Linguistics: EMNLP 2025}, pages 3172--3196, Suzhou, China. Association for Computational Linguistics.

\bibitem[{Khoja et~al.(2025)Khoja, Gupta, Fu, Roth, and Gupta}]{khoja-etal-2025-weaver}
Rohit Khoja, Devanshu Gupta, Yanjie Fu, Dan Roth, and Vivek Gupta. 2025.
\newblock \href {https://doi.org/10.18653/v1/2025.emnlp-main.1436} {Weaver: Interweaving {SQL} and {LLM} for table reasoning}.
\newblock In \emph{Proceedings of the 2025 Conference on Empirical Methods in Natural Language Processing}, pages 28282--28308, Suzhou, China. Association for Computational Linguistics.

\bibitem[{Lee et~al.(2025)Lee, Kim, Lee, Lee, and Kim}]{lee2024piece}
Wonjin Lee, Kyumin Kim, Sungjae Lee, Jihun Lee, and Kwang~In Kim. 2025.
\newblock \href {https://arxiv.org/abs/2412.07629} {Piece of table: A divide-and-conquer approach for selecting subtables in table question answering}.

\bibitem[{Li et~al.(2023)Li, Hui, Qu, Yang, Li, Li, Wang, Qin, Geng, Huo, Zhou, Ma, Li, Chang, Huang, Cheng, and Li}]{10.5555/3666122.3667957}
Jinyang Li, Binyuan Hui, Ge~Qu, Jiaxi Yang, Binhua Li, Bowen Li, Bailin Wang, Bowen Qin, Ruiying Geng, Nan Huo, Xuanhe Zhou, Chenhao Ma, Guoliang Li, Kevin~C.C. Chang, Fei Huang, Reynold Cheng, and Yongbin Li. 2023.
\newblock Can llm already serve as a database interface? a big bench for large-scale database grounded text-to-sqls.
\newblock In \emph{Proceedings of the 37th International Conference on Neural Information Processing Systems}, NIPS '23, Red Hook, NY, USA. Curran Associates Inc.

\bibitem[{Li et~al.(2025)Li, Tian, Chen, Ye, Ye, Wang, Wang, Fu, Chen, and Zhao}]{li-etal-2025-longtablebench}
Liyao Li, Jiaming Tian, Hao Chen, Wentao Ye, Chao Ye, Haobo Wang, Ningtao Wang, Xing Fu, Gang Chen, and Junbo Zhao. 2025.
\newblock \href {https://doi.org/10.18653/v1/2025.findings-emnlp.638} {{L}ong{T}able{B}ench: Benchmarking long-context table reasoning across real-world formats and domains}.
\newblock In \emph{Findings of the Association for Computational Linguistics: EMNLP 2025}, pages 11927--11965, Suzhou, China. Association for Computational Linguistics.

\bibitem[{Lin et~al.(2023)Lin, Blloshmi, Byrne, de~Gispert, and Iglesias}]{lin-etal-2023-inner}
Weizhe Lin, Rexhina Blloshmi, Bill Byrne, Adria de~Gispert, and Gonzalo Iglesias. 2023.
\newblock \href {https://doi.org/10.18653/v1/2023.acl-long.551} {An inner table retriever for robust table question answering}.
\newblock In \emph{Proceedings of the 61st Annual Meeting of the Association for Computational Linguistics (Volume 1: Long Papers)}, pages 9909--9926, Toronto, Canada. Association for Computational Linguistics.

\bibitem[{Liu et~al.(2022)Liu, Chen, Guo, Lin, Lou, and Zhang}]{liu2022tapex}
Qian Liu, Bei Chen, Jiaqi Guo, Zhenhua Lin, Jian-Guang Lou, and Dongmei Zhang. 2022.
\newblock Tapex: Table pre-training via learning a neural sql executor.
\newblock In \emph{International Conference on Learning Representations (ICLR)}.

\bibitem[{Lu et~al.(2025)Lu, Zhang, Fan, Fu, Chen, and Du}]{lu2025tablellmsurvey}
Weizheng Lu, Jing Zhang, Ju~Fan, Zihao Fu, Yueguo Chen, and Xiaoyong Du. 2025.
\newblock \href {https://doi.org/10.1007/s11704-024-40763-6} {Large language models for table processing: A survey}.
\newblock \emph{Frontiers of Computer Science}, 19(2).

\bibitem[{Mao et~al.(2026)Mao, Liu, Li, Cheng, Zhang, and Li}]{mao2026potablesystematicthinkingplanthenexecute}
Qingyang Mao, Qi~Liu, Zhi Li, Mingyue Cheng, Zheng Zhang, and Rui Li. 2026.
\newblock \href {https://arxiv.org/abs/2412.04272} {Potable: Towards systematic thinking via plan-then-execute stage reasoning on tables}.
\newblock \emph{Preprint}, arXiv:2412.04272.

\bibitem[{Nahid and Rafiei(2024{\natexlab{a}})}]{nahid-rafiei-2024-normtab}
Md~Mahadi~Hasan Nahid and Davood Rafiei. 2024{\natexlab{a}}.
\newblock \href {https://doi.org/10.18653/v1/2024.findings-emnlp.203} {{N}orm{T}ab: Improving symbolic reasoning in {LLM}s through tabular data normalization}.
\newblock In \emph{Findings of the Association for Computational Linguistics: EMNLP 2024}, pages 3569--3585, Miami, Florida, USA. Association for Computational Linguistics.

\bibitem[{Nahid and Rafiei(2024{\natexlab{b}})}]{nahid-rafiei-2024-tabsqlify}
Md~Mahadi~Hasan Nahid and Davood Rafiei. 2024{\natexlab{b}}.
\newblock \href {https://doi.org/10.18653/v1/2024.naacl-long.320} {{T}ab{SQL}ify: Enhancing reasoning capabilities of {LLM}s through table decomposition}.
\newblock In \emph{Proceedings of the 2024 Conference of the North American Chapter of the Association for Computational Linguistics: Human Language Technologies (Volume 1: Long Papers)}, pages 5725--5737, Mexico City, Mexico. Association for Computational Linguistics.

\bibitem[{Nahid and Rafiei(2025)}]{nahid2025prism}
Md~Mahadi~Hasan Nahid and Davood Rafiei. 2025.
\newblock \href {https://arxiv.org/abs/2510.14278} {Prism: Agentic retrieval with llms for multi-hop question answering}.

\bibitem[{Nahid et~al.(2026)Nahid, Rafiei, Zhang, and Zhang}]{nahid-etal-2026-rethinking}
Md~Mahadi~Hasan Nahid, Davood Rafiei, Weiwei Zhang, and Yong Zhang. 2026.
\newblock \href {https://doi.org/10.18653/v1/2026.findings-eacl.236} {Rethinking schema linking: A context-aware bidirectional retrieval approach for text-to-{SQL}}.
\newblock In \emph{Findings of the {A}ssociation for {C}omputational {L}inguistics: {EACL} 2026}, pages 4516--4546, Rabat, Morocco. Association for Computational Linguistics.

\bibitem[{Pasupat and Liang(2015)}]{pasupat-liang-2015-compositional}
Panupong Pasupat and Percy Liang. 2015.
\newblock \href {https://doi.org/10.3115/v1/P15-1142} {Compositional semantic parsing on semi-structured tables}.
\newblock In \emph{Proceedings of the 53rd Annual Meeting of the Association for Computational Linguistics and the 7th International Joint Conference on Natural Language Processing (Volume 1: Long Papers)}, pages 1470--1480, Beijing, China. Association for Computational Linguistics.

\bibitem[{Patnaik et~al.(2024)Patnaik, Changwal, Aggarwal, Bhatia, Kumar, and Krishnamurthy}]{patnaik2024cabinet}
Sohan Patnaik, Heril Changwal, Milan Aggarwal, Sumit Bhatia, Yaman Kumar, and Balaji Krishnamurthy. 2024.
\newblock \href {https://openreview.net/forum?id=SQrHpTllXa} {{CABINET}: Content relevance-based noise reduction for table question answering}.
\newblock In \emph{The Twelfth International Conference on Learning Representations}.

\bibitem[{Tyagi et~al.(2025)Tyagi, Gupta, and Bouri}]{tyagi2024agentictableqa}
Rishit Tyagi, Mohit Gupta, and Rahul Bouri. 2025.
\newblock \href {https://arxiv.org/abs/2509.09234} {Agentic large language models for question answering over tabular data}.

\bibitem[{Wang et~al.(2025{\natexlab{a}})Wang, Zheng, Tang, Lin, Cao, Wang, Cai, and Wang}]{wang2025needleinatable}
Lanrui Wang, Mingyu Zheng, Hongyin Tang, Zheng Lin, Yanan Cao, Jingang Wang, Xunliang Cai, and Weiping Wang. 2025{\natexlab{a}}.
\newblock Needleinatable: Exploring long-context capability of large language models towards long-structured tables.
\newblock In \emph{39th Conference on Neural Information Processing Systems (NeurIPS 2025)}.

\bibitem[{Wang et~al.(2025{\natexlab{b}})Wang, Gan, and Qi}]{wang2025tabsd}
Yuxiang Wang, Junhao Gan, and Jianzhong Qi. 2025{\natexlab{b}}.
\newblock \href {https://arxiv.org/abs/2502.13422} {Tabsd: Sql-based table decomposition for free-form table question answering}.

\bibitem[{Wang et~al.(2025{\natexlab{c}})Wang, Zhang, Nie, and Mao}]{wang2024tabaf}
Zhongyuan Wang, Richong Zhang, Zhijie Nie, and Hangyu Mao. 2025{\natexlab{c}}.
\newblock \href {https://arxiv.org/abs/2503.12345} {General table question answering via answer-formula joint generation}.

\bibitem[{Wang et~al.(2024)Wang, Zhang, Li, Eisenschlos, Perot, Wang, Miculicich, Fujii, Shang, Lee, and Pfister}]{wang2024chainoftable}
Zilong Wang, Hao Zhang, Chun-Liang Li, Julian~Martin Eisenschlos, Vincent Perot, Zifeng Wang, Lesly Miculicich, Yasuhisa Fujii, Jingbo Shang, Chen-Yu Lee, and Tomas Pfister. 2024.
\newblock \href {https://openreview.net/forum?id=4L0xnS4GQM} {Chain-of-table: Evolving tables in the reasoning chain for table understanding}.
\newblock In \emph{The Twelfth International Conference on Learning Representations}.

\bibitem[{Wu et~al.(2025)Wu, Yang, Chai, Zhang, Liu, Du, Liang, Shu, Cheng, Sun et~al.}]{wu2025tablebench}
Xianjie Wu, Jian Yang, Linzheng Chai, Ge~Zhang, Jiaheng Liu, Xeron Du, Di~Liang, Daixin Shu, Xianfu Cheng, Tianzhen Sun, and 1 others. 2025.
\newblock Tablebench: A comprehensive and complex benchmark for table question answering.
\newblock In \emph{Proceedings of the AAAI Conference on Artificial Intelligence}, volume~39, pages 25497--25506.

\bibitem[{Yang et~al.(2025)Yang, Du, Zhang, Du, Chen, Duan, and Zhao}]{yang2025triples}
Zhen Yang, Ziwei Du, Minghan Zhang, Wei Du, Jie Chen, Zhen Duan, and Shu Zhao. 2025.
\newblock \href {https://openreview.net/forum?id=UwcZEoNP19} {Triples as the key: Structuring makes decomposition and verification easier in {LLM}-based table{QA}}.
\newblock In \emph{The Thirteenth International Conference on Learning Representations}.

\bibitem[{Ye et~al.(2023)Ye, Hui, Yang, Li, Huang, and Li}]{10.1145/3539618.3591708}
Yunhu Ye, Binyuan Hui, Min Yang, Binhua Li, Fei Huang, and Yongbin Li. 2023.
\newblock \href {https://doi.org/10.1145/3539618.3591708} {Large language models are versatile decomposers: Decomposing evidence and questions for table-based reasoning}.
\newblock In \emph{Proceedings of the 46th International ACM SIGIR Conference on Research and Development in Information Retrieval}, SIGIR '23, page 174–184, New York, NY, USA. Association for Computing Machinery.

\bibitem[{Yu et~al.(2018)Yu, Zhang, Yang, Yasunaga, Wang, Li, Ma, Li, Yao, Roman, Zhang, and Radev}]{yu-etal-2018-spider}
Tao Yu, Rui Zhang, Kai Yang, Michihiro Yasunaga, Dongxu Wang, Zifan Li, James Ma, Irene Li, Qingning Yao, Shanelle Roman, Zilin Zhang, and Dragomir Radev. 2018.
\newblock \href {https://doi.org/10.18653/v1/D18-1425} {{S}pider: A large-scale human-labeled dataset for complex and cross-domain semantic parsing and text-to-{SQL} task}.
\newblock In \emph{Proceedings of the 2018 Conference on Empirical Methods in Natural Language Processing}, pages 3911--3921, Brussels, Belgium. Association for Computational Linguistics.

\bibitem[{Yu et~al.(2025)Yu, Jian, and Chen}]{yu-etal-2025-tablerag}
Xiaohan Yu, Pu~Jian, and Chong Chen. 2025.
\newblock \href {https://doi.org/10.18653/v1/2025.emnlp-main.710} {{T}able{RAG}: A retrieval augmented generation framework for heterogeneous document reasoning}.
\newblock In \emph{Proceedings of the 2025 Conference on Empirical Methods in Natural Language Processing}, pages 14063--14082, Suzhou, China. Association for Computational Linguistics.

\bibitem[{Zhang et~al.(2025{\natexlab{a}})Zhang, Chen, and Zhang}]{zhang2025tablemoeneurosymbolicroutingstructured}
Junwen Zhang, Pu~Chen, and Yin Zhang. 2025{\natexlab{a}}.
\newblock \href {https://arxiv.org/abs/2506.21393} {Tablemoe: Neuro-symbolic routing for structured expert reasoning in multimodal table understanding}.
\newblock \emph{Preprint}, arXiv:2506.21393.

\bibitem[{Zhang et~al.(2024{\natexlab{a}})Zhang, Luu, and Zhao}]{zhang-etal-2024-syntqa}
Siyue Zhang, Anh~Tuan Luu, and Chen Zhao. 2024{\natexlab{a}}.
\newblock \href {https://doi.org/10.18653/v1/2024.findings-emnlp.131} {{S}yn{TQA}: Synergistic table-based question answering via mixture of text-to-{SQL} and {E}2{E} {TQA}}.
\newblock In \emph{Findings of the Association for Computational Linguistics: EMNLP 2024}, pages 2352--2364, Miami, Florida, USA. Association for Computational Linguistics.

\bibitem[{Zhang et~al.(2025{\natexlab{b}})Zhang, Wang, Xu, Zhu, and Che}]{zhang-etal-2025-rot}
Xuanliang Zhang, Dingzirui Wang, Keyan Xu, Qingfu Zhu, and Wanxiang Che. 2025{\natexlab{b}}.
\newblock \href {https://doi.org/10.18653/v1/2025.emnlp-main.29} {{R}o{T}: Enhancing table reasoning with iterative row-wise traversals}.
\newblock In \emph{Proceedings of the 2025 Conference on Empirical Methods in Natural Language Processing}, pages 559--579, Suzhou, China. Association for Computational Linguistics.

\bibitem[{Zhang et~al.(2024{\natexlab{b}})Zhang, Henkel, Floratou, Cahoon, Deep, and Patel}]{10.14778/3659437.3659452}
Yunjia Zhang, Jordan Henkel, Avrilia Floratou, Joyce Cahoon, Shaleen Deep, and Jignesh~M. Patel. 2024{\natexlab{b}}.
\newblock \href {https://doi.org/10.14778/3659437.3659452} {Reactable: Enhancing react for table question answering}.
\newblock \emph{Proc. VLDB Endow.}, 17(8):1981–1994.

\bibitem[{Zhang et~al.(2024{\natexlab{c}})Zhang, Gao, and Lou}]{zhang-etal-2024-e5}
Zhehao Zhang, Yan Gao, and Jian-Guang Lou. 2024{\natexlab{c}}.
\newblock \href {https://doi.org/10.18653/v1/2024.naacl-long.68} {$e^5$: Zero-shot hierarchical table analysis using augmented {LLM}s via explain, extract, execute, exhibit and extrapolate}.
\newblock In \emph{Proceedings of the 2024 Conference of the North American Chapter of the Association for Computational Linguistics: Human Language Technologies (Volume 1: Long Papers)}, pages 1244--1258, Mexico City, Mexico. Association for Computational Linguistics.

\bibitem[{Zhou et~al.(2025)Zhou, Ma, Friedrich, and Mesgar}]{zhou2025tqasurvey}
Wei Zhou, Bolei Ma, Annemarie Friedrich, and Mohsen Mesgar. 2025.
\newblock \href {https://arxiv.org/abs/2510.09671} {Table question answering in the era of large language models: A survey}.

\end{thebibliography}

\clearpage

\appendix


\section{Additional Results}
\label{app:additional_results}

This appendix reports additional experiments and diagnostics conducted to
further evaluate the effectiveness, robustness, and efficiency of
\textsc{PARTAB}. Unless stated otherwise, \textsc{PARTAB} uses the default
LLM-based selection strategy and the same evaluation metrics as the main
paper.


\paragraph{Budget-Matched Full-Table Control.}
\label{app:budget_matched}

To distinguish the contribution of partition-aware reasoning from the
additional inference budget, we compare \textsc{PARTAB} with a full-table
baseline that uses the same number of LLM calls. The budget-matched baseline
repeatedly reasons over the complete serialized table but does not perform
semantic grouping or hierarchical partition selection.

\begin{table}[ht]
\centering
\small
\resizebox{\linewidth}{!}{
\begin{tabular}{l|c|c}
\toprule
{Method}
& {LLM Calls}
& {WikiTQ EM} \\
\midrule
Full-table (Budget-matched) 
& 5
& 64.61 \\

\textsc{PARTAB}
& 5
& \textbf{79.31} \\
\bottomrule
\end{tabular}
}
\caption{
Budget-matched comparison between repeated full-table prompting and
\textsc{PARTAB}.
}
\label{tab:budget_matched}
\end{table}

As shown in Table~\ref{tab:budget_matched}, the five-call full-table baseline
obtains 64.61 exact match on the evaluated WikiTQ subset, whereas
\textsc{PARTAB} obtains 79.31, an improvement of 14.70 points. Thus,
repeatedly prompting the model with the full table does not reproduce the
benefits of explicitly constructing a localized evidence state.

This control indicates that the performance improvement is attributable
primarily to semantic grouping, schema-level routing, and row-level
partition selection rather than merely to additional LLM interactions.

\begin{table}[h]
\centering
\small
\resizebox{\linewidth}{!}{
\begin{tabular}{l|c|c}
\toprule
{Dataset}
& {Fallback Rate (\%)}
& {Evidence Recall (\%)} \\
\midrule
WikiTQ
& 2.70
& 87.0$^{\dagger}$ \\

TabFact
& 1.90
& -- \\

TableBench
& 0.24
& -- \\
\bottomrule
\end{tabular}
}
\caption{
Selection stability across the evaluation datasets.
$^{\dagger}$Evidence recall is based on manual annotation of 100 WikiTQ
examples.
}
\label{tab:fallback_evidence}
\end{table}

\paragraph{Selection Stability and Fallback Frequency.}
\label{app:fallback_stability}

The LLM-based Part Selector may occasionally return an empty or malformed
selection. In such cases, \textsc{PARTAB} invokes the deterministic
\textsc{BasicSelect} fallback to preserve evidence coverage.

Table~\ref{tab:fallback_evidence} reports the empirical fallback frequency.
The fallback is activated for only 2.7\% of WikiTQ examples, 1.9\% of
TabFact examples, and 0.24\% of TableBench examples. Most fallback cases
are caused by malformed or unparsable structured outputs rather than an
explicit inability to identify relevant evidence.

These low rates show that the LLM-guided selection procedure is generally
stable. The fallback primarily serves as a robustness safeguard for rare
generation or parsing failures and does not drive the reported performance.

\paragraph{Evidence Coverage.}
\label{app:evidence_coverage}

The evaluated datasets do not provide gold cell-, row-, or column-level
evidence annotations. We therefore manually inspect 100 WikiTQ examples and
determine whether the partitions selected by \textsc{PARTAB} contain all
evidence necessary to answer the corresponding question.

The selected partitions achieve an evidence recall of 87\%. This result
indicates that the compact contexts constructed by \textsc{PARTAB} contain
sufficient evidence in most cases. The remaining failures primarily involve
aggregation-heavy questions requiring broad row coverage or cases where a
required row occurs in an unselected partition.

This analysis separates evidence-localization failures from downstream
execution errors. In particular, an incorrect final answer does not
necessarily imply that partition selection failed, because the answer
executor may still make a reasoning or computation error after receiving
the correct evidence.







\section{Additional Analysis}
\label{app:additional_analysis}


\begin{table*}[t]
\centering
\small

\resizebox{.85\linewidth}{!}{
\setlength{\tabcolsep}{12pt}
\renewcommand{\arraystretch}{1.08}

\begin{tabular}{l|c|c|c}
\toprule
{Method}
& {LLM Calls}
& {Avg. Tokens / Query}
& {TabFact Acc.} \\
\midrule

Full-table prompting
& 1
& 2.1k
& 73.22 \\

Chain-of-Table
& $\leq 25$
& 13.2k
& 85.09 \\

TableMaster
& 11
& 3.3k
& 90.12 \\

\midrule
\textsc{PARTAB} -- Basic
& 4
& 2.1k
& 87.88 \\

\textsc{PARTAB} -- Similarity
& 4
& \textbf{1.6k}
& 85.87 \\

\textsc{PARTAB} -- LLM-based
& 5
& 2.3k
& \textbf{90.48} \\

\bottomrule
\end{tabular}
}
\caption{
Accuracy--efficiency comparison on TabFact. Token counts cover the complete
per-query pipeline. Bold values indicate the lowest token usage and the
highest accuracy, respectively.
}
\label{tab:tabfact_efficiency}
\end{table*}

\paragraph{Accuracy--Efficiency Trade-off.}
\label{app:efficiency_analysis}

We further compare the inference efficiency of \textsc{PARTAB} with
full-table prompting and recent multi-stage table reasoning systems.
Table~\ref{tab:tabfact_efficiency} reports the number of LLM calls, average
tokens processed per query, and TabFact accuracy.

The default LLM-based configuration of \textsc{PARTAB} uses five LLM calls
and approximately 2.3k tokens per query. In comparison, Chain-of-Table
requires up to 25 calls and approximately 13.2k tokens, while TableMaster
uses 11 calls and approximately 3.3k tokens. Despite using substantially
fewer calls, \textsc{PARTAB} obtains the highest accuracy among the compared
methods.

The similarity-based variant is the most token-efficient configuration,
using approximately 1.6k tokens per query. However, its lower accuracy shows
that reducing context is beneficial only when the selected evidence is
semantically adequate. Aggressive context reduction based on lexical
similarity can omit evidence required for compositional reasoning.







\begin{table}[h]
\centering
\small
\resizebox{0.95\linewidth}{!}{
\setlength{\tabcolsep}{6pt}

\begin{tabular}{l|c|c}
\toprule
Method & Avg. Selected Parts ↓ & WikiTQ (EM) \\
\midrule
Full Table               & 17.56 & 59.43 \\
Basic Selection          & 6.50  & 74.71 \\
Similarity-based         & 5.36  & 67.25 \\
LLM-based                & \textbf{3.88}  & \textbf{79.31} \\
\bottomrule
\end{tabular}
}
\caption{
Partition reduction efficiency on WikiTableQuestions. We report the average number of selected parts and Exact Match (EM) accuracy.
}
\label{tab:partition_efficiency}
\end{table}

\paragraph{Partition Reduction Efficiency.}

Table~\ref{tab:partition_efficiency} shows that PARTAB significantly reduces the amount of context required for reasoning while improving performance. Compared to full-table prompting, all partition-based methods select substantially fewer parts, leading to more focused reasoning.

Among them, LLM-based selection achieves the best trade-off, reducing the average number of selected parts to 3.88 while achieving the highest accuracy (79.31 EM). This indicates that effective partition selection not only improves efficiency but also enhances reasoning quality by filtering out irrelevant context. In contrast, similarity-based selection reduces the number of parts but suffers from lower accuracy, highlighting that aggressive reduction without semantic understanding can harm performance.


\paragraph{Dataset Structural Statistics.}
\label{app:dataset_statistics}

The evaluated datasets differ substantially in table size and structure, as
shown in Table~\ref{tab:dataset_statistics}. WikiTQ contains the longest
tables, including examples with up to 517 rows. TableBench contains examples
with up to 176 rows, whereas TabFact tables contain at most 47 rows.

All three benchmarks also contain wide schemas, with maximum column counts
between 20 and 21. These statistics motivate partitioning along both table
dimensions. Long tables require row-level evidence localization, while wide
tables require schema-level routing to prevent irrelevant columns from
diluting the executor's attention.

\begin{table*}[t]
\centering
\small
\resizebox{.85\linewidth}{!}{
\setlength{\tabcolsep}{12pt}
\renewcommand{\arraystretch}{1.08}

\begin{tabular}{l|ccc | ccc}
\toprule
\multirow{2}{*}{Dataset}
& \multicolumn{3}{c}{Number of Rows}
& \multicolumn{3}{c}{Number of Columns} \\
\cmidrule(lr){2-4}
\cmidrule(lr){5-7}
& Min
& Avg
& Max
& Min
& Avg
& Max \\
\midrule
WikiTableQuestions
& 5
& 25.65
& 517
& 3
& 6.39
& 21 \\

TabFact
& 5
& 13.41
& 47
& 5
& 6.29
& 20 \\

TableBench
& 6
& 16.75
& 176
& 4
& 6.64
& 20 \\
\bottomrule
\end{tabular}
}
\caption{Structural statistics of the evaluation datasets.}
\label{tab:dataset_statistics}
\end{table*}

\section{Discussions}

\paragraph{When Partition-Aware Reasoning Helps.}
\label{app:when_partitioning_helps}

The additional experiments provide a more precise characterization of when
partition-aware reasoning is beneficial. The largest improvements occur on
long or wide tables, where the evidence required by a question occupies only
a small portion of the complete table. In such settings, full-table prompting exposes the model to many irrelevant values, increasing attention dilution and making it harder to localize and reason over the relevant evidence.

Partitioning is particularly effective for lookup, comparison, and fact
verification questions. These tasks usually depend on a localized set of
attributes and rows, making it possible to construct a compact evidence
state without losing task-relevant information. The large gains on
TabFact-Hard across all three model backbones provide strong evidence for
this behavior.

Partition-aware reasoning is also useful when table cells contain noisy or
free-form text that is difficult to filter using exact symbolic predicates.
Semantic grouping and LLM-based selection can identify conceptually relevant
attributes even when the question and table use different surface forms.

\paragraph{Relationship to Full-Table and Single-View Reasoning.}
\label{app:reasoning_distinction}

The benefit of \textsc{PARTAB} is not simply that it passes fewer tokens to
the answer executor. A pruning method can also reduce context length, but
usually constructs a single reduced table. If that reduced view omits a
required column or row, the missing evidence cannot be recovered during
reasoning.

In contrast, \textsc{PARTAB} preserves a collection of complementary,
row-linked table parts. Semantic group selection first identifies the
relevant schema subspace, after which part selection identifies relevant row
chunks within that subspace. The universal \texttt{row\_id} key allows
evidence from different semantic groups to be aligned during answer
execution.

Thus, partitioning acts as a reasoning control mechanism rather than only a
preprocessing operation. The model reasons over multiple structured table
regions while preserving their relational correspondence instead of
receiving either the complete table or a single flattened sub-table.

\paragraph{Relationship to Text-to-SQL Reasoning.}
\label{app:text_to_sql_discussion}

Text-to-SQL methods are highly effective when a question can be translated
into precise filtering, joining, sorting, or aggregation operations.
Executing the generated query in a database engine also provides reliable
numerical computation.

However, the evaluated benchmarks include questions requiring semantic
verification, implicit comparison, and interpretation of noisy or
free-form textual cells. These operations are not always naturally
expressible as exact SQL predicates. Text-to-SQL systems may therefore fail
during predicate construction even when the necessary evidence is present.

\textsc{PARTAB} addresses a complementary bottleneck: how to localize and
structure evidence before semantic reasoning. It does not replace symbolic
execution. For completeness-sensitive aggregation questions, a promising
extension would combine partition-aware semantic localization with
deterministic SQL or program execution.

\paragraph{Remaining Limitations.}
\label{app:additional_limitations}

The additional analyses also clarify several limitations of the current
framework. First, part selection remains the dominant source of error.
Although the selected partitions obtain 87\% evidence recall on the manually
annotated WikiTQ sample, the remaining selection failures can directly
prevent correct downstream reasoning.

Second, localized evidence selection is less suitable for questions
requiring exhaustive aggregation over most or all table rows. Selecting only
a subset of partitions can violate the completeness requirement of counting,
global extrema, or large-scale aggregation operations.

Third, the effectiveness of the pipeline remains dependent on the
instruction-following and structured-output capabilities of the selected
LLM backbone. 


A promising extension is an adaptive global--local controller that first
identifies whether the question requires localized semantic evidence or
global table completeness. Localized questions could use the current
partition-aware pipeline, while completeness-sensitive questions could be
routed to full-table or symbolic execution.



\section{Extended Literature Review}



\paragraph{LLM-based Table Question Answering.}
Recent advances have significantly improved Table Question Answering (TableQA) using large language models. Surveys highlight rapid progress in LLM-based table understanding while identifying scalability and reasoning reliability as major challenges~\citep{zhou2025tqasurvey,lu2025tablellmsurvey,fang2024tabularllm, cao2026tablemasterrecipeadvancetable}. Direct prompting approaches enable flexible semantic reasoning but struggle with long tables due to attention limitations and incomplete evidence utilization~\citep{tyagi2024agentictableqa}. Earlier transformer-based models such as TAPAS demonstrated the effectiveness of pretraining for tabular reasoning, establishing foundations for modern LLM-based approaches~\citep{herzig2020tapas}.

\paragraph{Text-to-SQL and Symbolic Table Reasoning.}
A dominant paradigm formulates table reasoning as Text-to-SQL generation, translating natural language queries into executable programs evaluated by database engines~\citep{yu-etal-2018-spider, nahid-etal-2026-rethinking}. Benchmarks such as BIRD emphasize realistic database reasoning scenarios requiring accurate aggregation and efficient execution~\citep{10.5555/3666122.3667957}. Symbolic execution enables reliable numerical reasoning; however, these approaches depend on constructing structured predicates and often struggle when filtering conditions require semantic interpretation over noisy textual cells. Recent work integrating LLM planning with executable reasoning further highlights both the strengths and limitations of symbolic pipelines~\citep{wang2024tabaf,evkarpidi2025bridging, liu2022tapex}.

\paragraph{Decomposition and Context Reduction for Long Tables.}
To address scalability challenges, several works reduce reasoning complexity through decomposition strategies. These include question decomposition and table decomposition methods that divide large tables into smaller reasoning units. TabSQLify introduces SQL-guided sub-table extraction to improve reasoning efficiency on large tables~\citep{nahid-rafiei-2024-tabsqlify}, while TabSD extends decomposition to free-form noisy tables~\citep{wang2025tabsd}. Structured reasoning frameworks such as Tree-of-Table further organize hierarchical reasoning over tabular data~\citep{ji2025treeoftable}. Long-context benchmarks including LongTableBench and NeedleInATable demonstrate that reasoning performance degrades substantially as table size increases~\citep{li-etal-2025-longtablebench,wang2025needleinatable, wu2025tablebench}. 

Recent work has explored decomposition, retrieval, and hybrid execution for table reasoning. Methods such as ITR~\cite{lin-etal-2023-inner} and PieTa~\cite{lee2024piece} focus on retrieving or constructing relevant sub-tables to reduce long-table noise, while RoT~\cite{zhang-etal-2025-rot} performs row-wise reasoning to improve scalability. CABINET~\cite{patnaik2024cabinet} addresses noisy tables through content relevance weighting, and TABDSR~\cite{jiang-etal-2025-tabdsr} combines decomposition with program-based reasoning for numerical tasks. Hybrid frameworks such as ReAcTable~\cite{10.14778/3659437.3659452} and Weaver~\cite{khoja-etal-2025-weaver} integrate LLM reasoning with external tools, while E$^5$~\cite{zhang-etal-2024-e5} and TIDE~\cite{yang2025triples} introduce structured intermediate representations for improved reasoning and verification. In contrast, PARTAB treats partitioning as a reasoning control mechanism, combining semantic grouping with row-linked selection to enable structured reasoning over partitioned table views.

While decomposition improves evidence localization, restricting reasoning to selected subsets introduces a critical trade-off: required evidence may be omitted when tasks demand exhaustive global computation.

\paragraph{Hybrid Neural--Symbolic Table Reasoning.}
An emerging direction combines LLM reasoning with executable intermediate representations such as SQL queries, programs, or dataframe operations. Hybrid systems leverage LLMs for interpretation while relying on deterministic execution engines for numerical correctness and interpretability~\citep{wang2024tabaf,evkarpidi2025bridging, abhyankar-etal-2025-h}. Approaches such as SynTQA demonstrate complementary strengths between symbolic execution and end-to-end reasoning under different task conditions~\citep{zhang-etal-2024-syntqa, cheng2023binder}. Neuro-symbolic routing frameworks such as TableMoE further explore structured expert reasoning for tabular understanding~\citep{zhang2025tablemoeneurosymbolicroutingstructured, 10.1145/3539618.3591708}.

\paragraph{Retrieval-Augmented Reasoning over Structured Data.}
Retrieval-Augmented Generation (RAG) has recently been extended to structured and semi-structured data sources, including tables and databases~\citep{tyagi2024agentictableqa, nahid2025prism}. Table-aware retrieval methods index table regions or schema components and retrieve relevant fragments prior to reasoning, improving scalability by avoiding full-context processing. Recent work has explored retrieval-augmented approaches for scalable table reasoning. \citet{chen2024tablerag} introduce TableRAG, which enables large-scale table understanding by retrieving relevant table segments from million-token contexts. Similarly, \citet{yu-etal-2025-tablerag} propose a retrieval-augmented framework for heterogeneous document reasoning that combines text retrieval with SQL-based execution over tabular data. However, retrieval-based approaches inherit an important limitation: incomplete retrieval can produce confident but incorrect answers, particularly for aggregation or counting tasks requiring full-table access.

\paragraph{Limitations of Existing Approaches.}
Despite substantial progress, existing methods expose complementary trade-offs. Symbolic approaches provide reliable aggregation but depend on precise structured predicates and can struggle with noisy or free-form textual content. Direct LLM reasoning offers greater semantic flexibility but becomes less reliable as table size and irrelevant context increase. Decomposition and retrieval methods improve scalability by reducing context, yet may omit evidence required for completeness-sensitive operations.

These trade-offs suggest that effective table reasoning should distinguish \emph{semantic evidence localization} from \emph{global execution}. Rather than treating any single reasoning paradigm as universally sufficient, a more general design should coordinate localized semantic reasoning with exhaustive or symbolic computation when required. \textbf{PARTAB} addresses the first part of this problem by constructing structured, query-relevant evidence states over table partitions, while complementary global or symbolic execution remains an important direction for completeness-sensitive reasoning.



\section{LLM Usage}

All technical content, contributions, analyses, and implementations are entirely our own. Large language models were used only to improve clarity and readability, and generating charts and diagrams. LLMs were occasionally used for minor editing and code debugging, with all outputs reviewed and verified by the authors.

\section{Illustrative Examples and Prompt Template}
\label{app:example_prompts}

\begin{figure*}[ht]
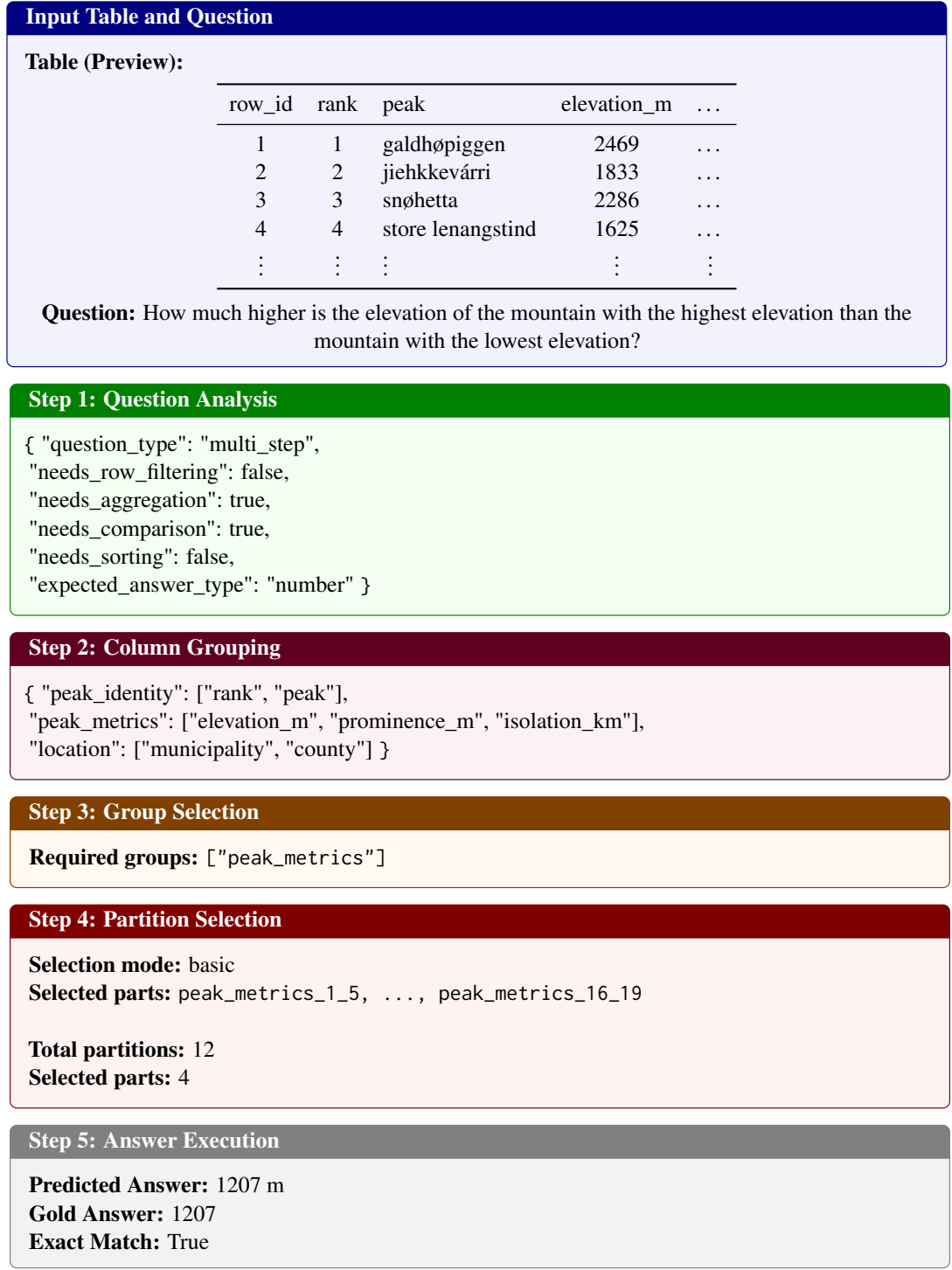

\centering
\resizebox{0.90\linewidth}{!}{
\begin{tcolorbox}[colframe=blue!50!black, colback=blue!5!white, coltitle=white, fonttitle=\bfseries, title=\textbf{Input Table and Question}]
\normalsize

\textbf{Table (Preview):}

\vspace{4pt}
\centering
\begin{tabular}{c c l c c}
\toprule
row\_id & rank & peak & elevation\_m & \dots \\
\midrule
1 & 1 & galdhøpiggen       & 2469 & \dots \\
2 & 2 & jiehkkevárri       & 1833 & \dots \\
3 & 3 & snøhetta           & 2286 & \dots \\
4 & 4 & store lenangstind  & 1625 & \dots \\
\vdots & \vdots & \vdots & \vdots & \vdots \\

\bottomrule
\end{tabular}

\vspace{6pt}
\textbf{Question:} How much higher is the elevation of the mountain with the highest elevation than the mountain with the lowest elevation?

\end{tcolorbox}
}
\vspace{6pt}

\resizebox{0.90\linewidth}{!}{
\begin{tcolorbox}[colframe=green!50!black, colback=green!5!white, fonttitle=\bfseries, title=\textbf{Step 1: Question Analysis}]
\normalsize

\texttt{\{}
"question\_type": "multi\_step", \\
"needs\_row\_filtering": false, \\
"needs\_aggregation": true, \\
"needs\_comparison": true, \\
"needs\_sorting": false, \\
"expected\_answer\_type": "number"
\texttt{\}}

\end{tcolorbox}
}

\vspace{6pt}

\resizebox{0.90\linewidth}{!}{
\begin{tcolorbox}[colframe=purple!50!black, colback=purple!5!white, fonttitle=\bfseries, title=\textbf{Step 2: Column Grouping}]
\normalsize

\texttt{\{}
"peak\_identity": ["rank", "peak"], \\
"peak\_metrics": ["elevation\_m", "prominence\_m", "isolation\_km"], \\
"location": ["municipality", "county"]
\texttt{\}}

\end{tcolorbox}
}

\vspace{6pt}

\resizebox{0.90\linewidth}{!}{
\begin{tcolorbox}[colframe=orange!50!black, colback=orange!5!white, fonttitle=\bfseries, title=\textbf{Step 3: Group Selection}]
\normalsize

\textbf{Required groups:} \texttt{["peak\_metrics"]}

\end{tcolorbox}
}

\vspace{6pt}

\resizebox{0.90\linewidth}{!}{
\begin{tcolorbox}[colframe=red!50!black, colback=red!5!white, fonttitle=\bfseries, title=\textbf{Step 4: Partition Selection}]
\normalsize

\textbf{Selection mode:} basic \\
\textbf{Selected parts:} \texttt{peak\_metrics\_1\_5, ..., peak\_metrics\_16\_19} \\

\textbf{Total partitions:} 12 \\
\textbf{Selected parts:} 4

\end{tcolorbox}
}

\vspace{6pt}

\resizebox{0.90\linewidth}{!}{
\begin{tcolorbox}[colframe=black!50, colback=gray!10, fonttitle=\bfseries, title=\textbf{Step 5: Answer Execution}]
\normalsize

\textbf{Predicted Answer:} 1207 m \\
\textbf{Gold Answer:} 1207 \\
\textbf{Exact Match:} True

\end{tcolorbox}
}

\caption{
Illustrative example of PARTAB on a multi-step numerical reasoning task. The table is truncated for clarity; ellipses ($\dots$) indicate omitted columns and rows. PARTAB isolates the relevant metric columns, selects a small subset of partitions, and correctly computes the difference between the highest and lowest elevations.
}
\label{fig:partab-example-numeric}

\end{figure*}

\begin{figure*}[ht]
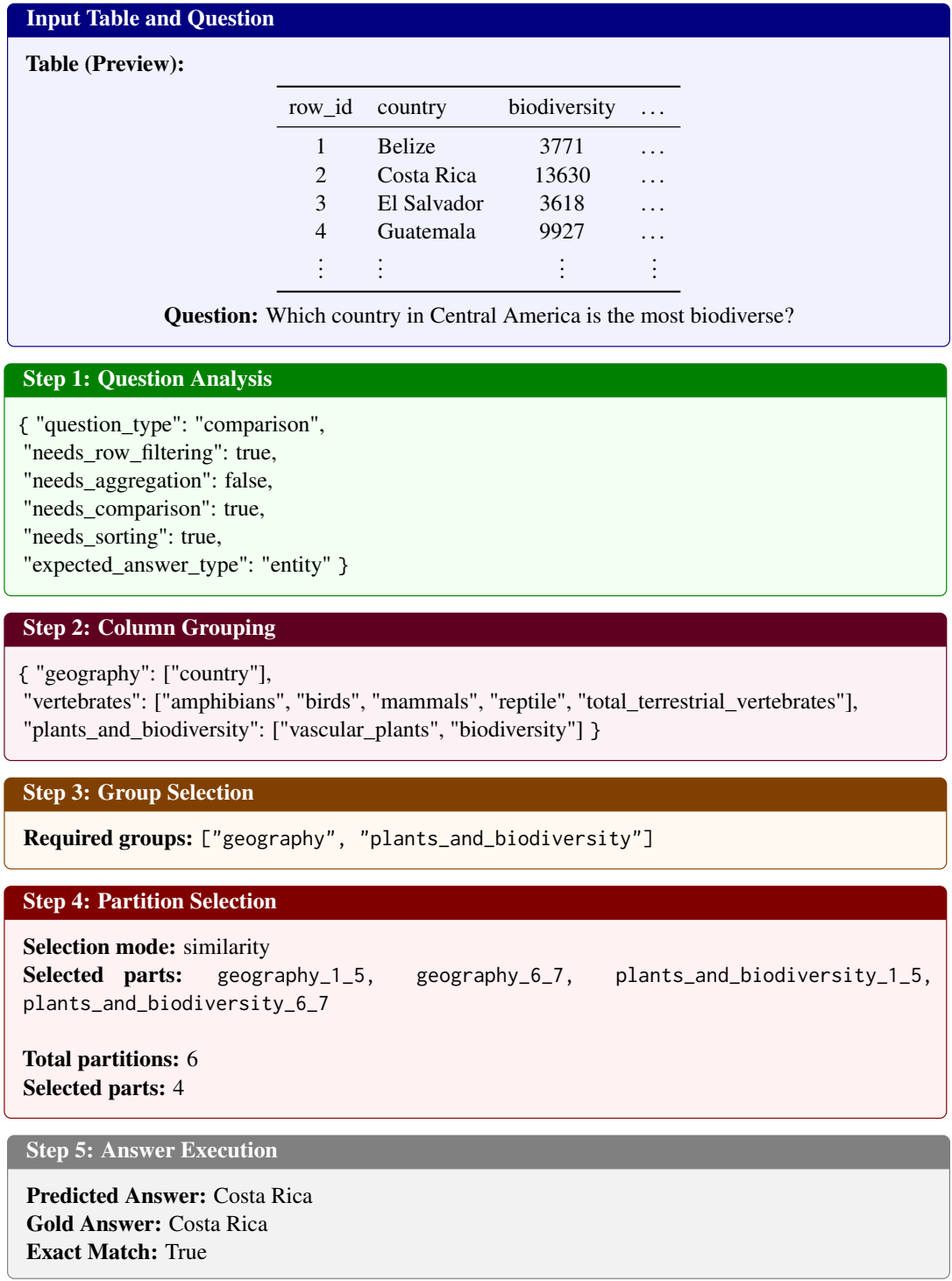

\centering
\resizebox{0.90\linewidth}{!}{
\begin{tcolorbox}[colframe=blue!50!black, colback=blue!5!white, coltitle=white, fonttitle=\bfseries, title=\textbf{Input Table and Question}]
\normalsize

\textbf{Table (Preview):}

\vspace{4pt}
\centering
\begin{tabular}{c l c c}
\toprule
row\_id & country & biodiversity & \dots \\
\midrule
1 & Belize      & 3771  & \dots \\
2 & Costa Rica  & 13630 & \dots \\
3 & El Salvador & 3618  & \dots \\
4 & Guatemala   & 9927  & \dots \\
\vdots & \vdots & \vdots & \vdots \\

\bottomrule
\end{tabular}

\vspace{6pt}
\textbf{Question:} Which country in Central America is the most biodiverse?

\end{tcolorbox}
}

\vspace{6pt}

\resizebox{0.90\linewidth}{!}{
\begin{tcolorbox}[colframe=green!50!black, colback=green!5!white, fonttitle=\bfseries, title=\textbf{Step 1: Question Analysis}]
\normalsize

\texttt{\{}
"question\_type": "comparison", \\
"needs\_row\_filtering": true, \\
"needs\_aggregation": false, \\
"needs\_comparison": true, \\
"needs\_sorting": true, \\
"expected\_answer\_type": "entity"
\texttt{\}}

\end{tcolorbox}
}
\vspace{6pt}

\resizebox{0.90\linewidth}{!}{
\begin{tcolorbox}[colframe=purple!50!black, colback=purple!5!white, fonttitle=\bfseries, title=\textbf{Step 2: Column Grouping}]
\normalsize

\texttt{\{}
"geography": ["country"], \\
"vertebrates": ["amphibians", "birds", "mammals", "reptile", "total\_terrestrial\_vertebrates"], \\
"plants\_and\_biodiversity": ["vascular\_plants", "biodiversity"]
\texttt{\}}

\end{tcolorbox}
}
\vspace{6pt}

\resizebox{0.90\linewidth}{!}{
\begin{tcolorbox}[colframe=orange!50!black, colback=orange!5!white, fonttitle=\bfseries, title=\textbf{Step 3: Group Selection}]
\normalsize

\textbf{Required groups:} \texttt{["geography", "plants\_and\_biodiversity"]}

\end{tcolorbox}
}
\vspace{6pt}

\resizebox{0.90\linewidth}{!}{
\begin{tcolorbox}[colframe=red!50!black, colback=red!5!white, fonttitle=\bfseries, title=\textbf{Step 4: Partition Selection}]
\normalsize

\textbf{Selection mode:} similarity \\
\textbf{Selected parts:} \texttt{geography\_1\_5, geography\_6\_7, plants\_and\_biodiversity\_1\_5, plants\_and\_biodiversity\_6\_7} \\

\textbf{Total partitions:} 6 \\
\textbf{Selected parts:} 4

\end{tcolorbox}
}
\vspace{6pt}

\resizebox{0.90\linewidth}{!}{
\begin{tcolorbox}[colframe=black!50, colback=gray!10, fonttitle=\bfseries, title=\textbf{Step 5: Answer Execution}]
\normalsize

\textbf{Predicted Answer:} Costa Rica \\
\textbf{Gold Answer:} Costa Rica \\
\textbf{Exact Match:} True

\end{tcolorbox}
}
\caption{
Illustrative example of PARTAB on a comparison-based reasoning task. The table is truncated for clarity; ellipses ($\dots$) indicate omitted columns and rows. PARTAB selects the relevant semantic groups and partitions, enabling accurate identification of the most biodiverse country.
}
\label{fig:partab-example-biodiversity}

\end{figure*}

\begin{figure*}[ht]
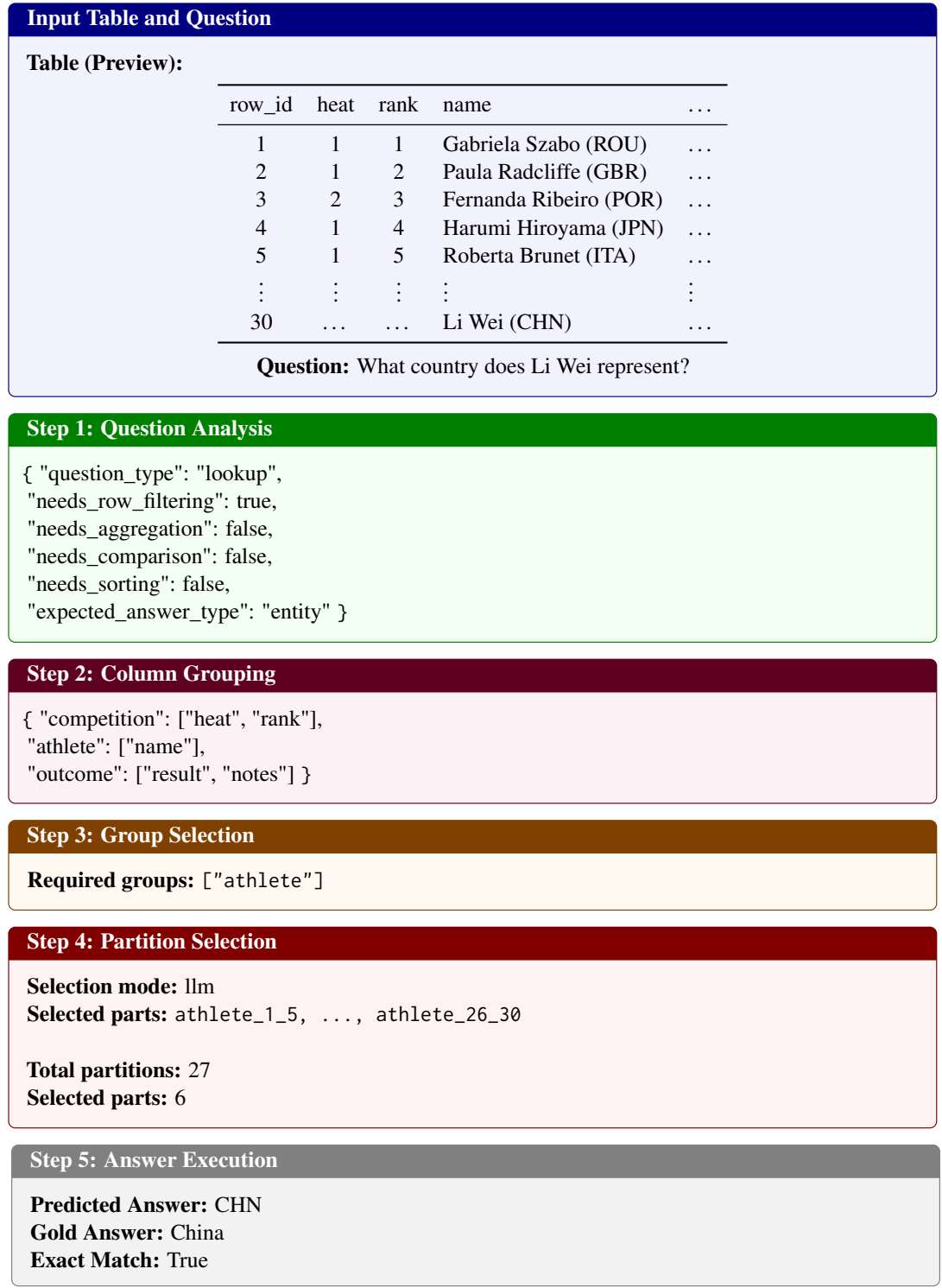

\centering
\resizebox{0.90\linewidth}{!}{
\begin{tcolorbox}[colframe=blue!50!black, colback=blue!5!white, coltitle=white, fonttitle=\bfseries, title=\textbf{Input Table and Question}]
\normalsize

\textbf{Table (Preview):}

\vspace{4pt}
\centering
\begin{tabular}{c c c l l}
\toprule
row\_id & heat & rank & name & \dots \\
\midrule
1 & 1 & 1 & Gabriela Szabo (ROU)   & \dots \\
2 & 1 & 2 & Paula Radcliffe (GBR)  & \dots \\
3 & 2 & 3 & Fernanda Ribeiro (POR) & \dots \\
4 & 1 & 4 & Harumi Hiroyama (JPN)  & \dots \\
5 & 1 & 5 & Roberta Brunet (ITA)   & \dots \\
\vdots & \vdots & \vdots & \vdots & \vdots \\
30 & \dots & \dots & Li Wei (CHN) & \dots \\
\bottomrule
\end{tabular}

\vspace{6pt}
\textbf{Question:} What country does Li Wei represent?

\end{tcolorbox}
}
\vspace{6pt}

\resizebox{0.90\linewidth}{!}{
\begin{tcolorbox}[colframe=green!50!black, colback=green!5!white, fonttitle=\bfseries, title=\textbf{Step 1: Question Analysis}]
\normalsize

\texttt{\{}
"question\_type": "lookup", \\
"needs\_row\_filtering": true, \\
"needs\_aggregation": false, \\
"needs\_comparison": false, \\
"needs\_sorting": false, \\
"expected\_answer\_type": "entity"
\texttt{\}}

\end{tcolorbox}
}
\vspace{6pt}

\resizebox{0.90\linewidth}{!}{
\begin{tcolorbox}[colframe=purple!50!black, colback=purple!5!white, fonttitle=\bfseries, title=\textbf{Step 2: Column Grouping}]
\normalsize

\texttt{\{}
"competition": ["heat", "rank"], \\
"athlete": ["name"], \\
"outcome": ["result", "notes"]
\texttt{\}}

\end{tcolorbox}
}
\vspace{6pt}

\resizebox{0.90\linewidth}{!}{
\begin{tcolorbox}[colframe=orange!50!black, colback=orange!5!white, fonttitle=\bfseries, title=\textbf{Step 3: Group Selection}]
\normalsize

\textbf{Required groups:} \texttt{["athlete"]}

\end{tcolorbox}
}
\vspace{6pt}

\resizebox{0.90\linewidth}{!}{
\begin{tcolorbox}[colframe=red!50!black, colback=red!5!white, fonttitle=\bfseries, title=\textbf{Step 4: Partition Selection}]
\normalsize

\textbf{Selection mode:} llm \\
\textbf{Selected parts:} \texttt{athlete\_1\_5, ..., athlete\_26\_30} \\

\textbf{Total partitions:} 27 \\
\textbf{Selected parts:} 6

\end{tcolorbox}
}
\vspace{6pt}

\resizebox{0.90\linewidth}{!}{
\begin{tcolorbox}[colframe=black!50, colback=gray!10, fonttitle=\bfseries, title=\textbf{Step 5: Answer Execution}]
\normalsize

\textbf{Predicted Answer:} CHN \\
\textbf{Gold Answer:} China \\
\textbf{Exact Match:} True

\end{tcolorbox}
}
\caption{
Illustrative example of PARTAB on a lookup-based reasoning task. The table is truncated for clarity; ellipses ($\dots$) indicate omitted rows and columns. PARTAB identifies the relevant row and extracts the country information from the athlete name field.
}
\label{fig:partab-example-lookup}

\end{figure*}

We present three illustrative examples to demonstrate how PARTAB operates across different reasoning scenarios, including numerical computation, comparison, and entity lookup. These examples highlight how partition-aware reasoning enables effective evidence localization and improves reasoning accuracy by focusing on relevant subsets of the table.
Figure~\ref{fig:partab-example-numeric}, \ref{fig:partab-example-biodiversity}, \ref{fig:partab-example-lookup}  illustrates a representative examples of three different categories. 

We also provide the detail prompt used to implement our method in Figure~\ref{fig:question-analyzer-prompt}, \ref{fig:column-grouping-prompt}, \ref{fig:group-selection-prompt}, \ref{fig:part-selection-prompt}, \ref{fig:answer-execution-prompt}. 


\clearpage


\begin{figure*}[ht]
\centering

\resizebox{1\linewidth}{!}{
\begin{tcolorbox}[colframe=blue!60!black, colback=blue!5!white, coltitle=white, fonttitle=\bfseries, fontupper=\itshape, title=\textbf{Question Analyzer Prompt}]
\small

You are a question analyzer for table reasoning. \\

You will receive a natural language question over a table. Your task is to infer the likely reasoning requirements of the question. In particular, determine the question type, whether row filtering is needed, whether aggregation, comparison, or sorting is required, and the expected answer type. \\

Question: \{QUESTION\} \\

Return STRICT JSON only in the following format: \\

\texttt{\{ \\
\ \ "question\_type": "lookup", \\
\ \ "needs\_row\_filtering": true, \\
\ \ "needs\_aggregation": false, \\
\ \ "needs\_comparison": false, \\
\ \ "needs\_sorting": false, \\
\ \ "expected\_answer\_type": "entity" \\
\}} \\

Allowed values for \texttt{question\_type}: \\
\texttt{lookup, aggregation, comparison, counting, sorting, multi\_step, other} \\

Allowed values for \texttt{expected\_answer\_type}: \\
\texttt{entity, number, boolean, text, list, other} \\

Only output a valid JSON object. Do not include \texttt{```json} or any additional explanation.
\end{tcolorbox}
}
\caption{Question Analyzer Prompt Template used to infer reasoning requirements from a natural language question.}
\label{fig:question-analyzer-prompt}

\end{figure*}


\begin{figure*}[ht]
\centering
\resizebox{1\linewidth}{!}{
\begin{tcolorbox}[colframe=blue!60!black, colback=blue!5!white, coltitle=white, fonttitle=\bfseries, fontupper=\itshape, title=\textbf{Column Grouping Prompt}]
\small

You are designing semantic column groups for partition-aware reasoning over a table. \\

You will receive: \\
- The question \\
- The table schema \\
- A few sample rows from the table \\

Your goal is to group the table columns into a small number of semantically coherent groups that support partition-aware reasoning. The universal linking key is \texttt{row\_id}. Do not include \texttt{row\_id} in any semantic group. Each non-key column must appear in exactly one group. Prefer 2 to 5 groups when possible, and use short, meaningful group names. \\

Question: \{QUESTION\} \\
Table schema: \{SCHEMA\} \\
Sample rows: \{TABLE\_PREVIEW\} \\

Return STRICT JSON only in the following format: \\

\texttt{\{ \\
\ \ "column\_groups": \{ \\
\ \ \ \ "group\_name\_1": ["col1", "col2"], \\
\ \ \ \ "group\_name\_2": ["col3", "col4"] \\
\ \ \}, \\
\ \ "key\_columns": ["row\_id"] \\
\}} \\

Only output a valid JSON object. Do not include \texttt{```json} or any additional explanation.
\end{tcolorbox}
}
\caption{Column Grouping Prompt Template used to construct semantically coherent column groups for partition-aware table reasoning.}
\label{fig:column-grouping-prompt}
\end{figure*}


\begin{figure*}[ht]
\centering
\resizebox{1\linewidth}{!}{
\begin{tcolorbox}[colframe=blue!60!black, colback=blue!5!white, coltitle=white, fonttitle=\bfseries, fontupper=\itshape, title=\textbf{Group Selection Prompt}]
\small 

You are selecting the minimum required column groups needed to answer a table question. \\

You will receive: \\
- The original question \\
- The question analysis output \\
- The available column groups \\

Your goal is to select the minimum set of column groups required to answer the question. If the question requires filtering, lookup, comparison, aggregation, or sorting, include the relevant groups. Return only valid group names from the provided list. \\

Question: \{QUESTION\} \\
Question analysis: \{QUESTION\_ANALYSIS\} \\
Available column groups: \{COLUMN\_GROUPS\} \\

Return STRICT JSON only in the following format: \\

\texttt{\{ \\
\ \ "required\_groups": ["group1", "group2"] \\
\}} \\

Only output a valid JSON object. Do not include \texttt{```json} or any additional explanation.
\end{tcolorbox}
}
\caption{Group Selection Prompt Template used to identify the minimum required semantic groups for answering a question.}
\label{fig:group-selection-prompt}
\end{figure*}


\begin{figure*}[ht]
\centering
\resizebox{1\linewidth}{!}{
\begin{tcolorbox}[colframe=blue!60!black, colback=blue!5!white, coltitle=white, fonttitle=\bfseries, fontupper=\itshape, title=\textbf{LLM-based Part Selection Prompt}]
\small 

You are selecting the minimum required table parts needed to answer a question. \\

You will receive: \\
- The original question \\
- A list of candidate table parts, where each part includes its part name, group name, row range, columns, and table content \\

Your goal is to select the minimum set of part names needed to answer the question. Only return valid part names from the provided candidate list. \\

Question: \{QUESTION\} \\
Candidate table parts: \{PARTS\} \\

Return STRICT JSON only in the following format: \\

\texttt{\{ \\
\ \ "required\_part\_names": ["part1", "part2"] \\
\}} \\

Only output a valid JSON object. Do not include \texttt{```json} or any additional explanation.
\end{tcolorbox}
}
\caption{LLM-based Part Selection Prompt Template used to select the minimum required table parts from the candidate partition set.}
\label{fig:part-selection-prompt}
\end{figure*}











\begin{figure*}[ht]
\centering
\resizebox{\linewidth}{!}{
\begin{tcolorbox}[colframe=blue!60!black, colback=blue!5!white, coltitle=white, fonttitle=\bfseries, fontupper=\itshape, title=\textbf{Answer Execution Prompt}]
\small 

You are a Table Question Answering assistant. \\

Answer the question using ONLY the provided table parts. The linking key across parts is \texttt{row\_id}. Do not use outside knowledge. Do not explain your reasoning. If the answer cannot be found from the table parts, return exactly: \texttt{I cannot find it.} \\

Question: \{QUESTION\} \\
Table parts: \{CONTEXT\} \\

Return only the final short answer.
\end{tcolorbox}
}
\caption{Answer Execution Prompt Template used to generate the final answer from the selected table parts.}
\label{fig:answer-execution-prompt}
\end{figure*}


\end{document}